\documentclass[runningheads]{llncs}
\usepackage{graphicx}
\usepackage{comment}
\usepackage{amsmath,amssymb} 
\usepackage{color}

\usepackage{wrapfig}
\usepackage{multirow}
\usepackage{caption}
\usepackage{graphicx}
\usepackage{subfigure}
\usepackage{algorithm}
\usepackage{algorithmic}
\usepackage{xcolor}
\usepackage{xspace}
\usepackage{algorithm}
\usepackage{algorithmic}
\usepackage{listings}
\usepackage{booktabs} 
\usepackage{soul}

\newcommand{\halffigurelength}{0.45}
\newcommand{\thirdfigurelength}{0.31}
\newcommand{\fourthfigurelength}{0.24}
\newcommand{\scaletable}{0.75}

\begin{document}
\pagestyle{headings}
\mainmatter

\title{Hashing-based Non-Maximum Suppression for Crowded Object Detection} 

\author{Jianfeng Wang, Xi Yin, Lijuan Wang, Lei Zhang}
\institute{Microsoft\\
\{jianfw, xiyin1, lijuanw, leizhang\}@microsoft.com}

\maketitle

\begin{abstract}
	In this paper, we propose an algorithm, named hashing-based non-maximum suppression  (HNMS) to efficiently suppress the non-maximum boxes for object detection.	Non-maximum suppression (NMS) is an essential component to suppress the boxes at closely located locations with similar shapes. The time cost tends to be huge when the number of boxes becomes large, especially for crowded scenes. The basic idea of HNMS is to firstly map each box to a discrete code (hash cell) and then remove the boxes with lower confidences if they are in the same cell. Considering the intersection-over-union (IoU) as the metric, we propose a simple yet effective hashing algorithm, named IoUHash, which guarantees that the boxes within the same cell are close enough by a lower IoU bound. For two-stage detectors, we replace NMS in region proposal network with HNMS, and observe significant speed-up with comparable accuracy. For one-stage detectors, HNMS is used as a pre-filter to speed up the suppression with a large margin. Extensive experiments are conducted on CARPK, SKU-110K, CrowdHuman datasets to demonstrate the efficiency and effectiveness of HNMS.
	Code is released at \url{https://github.com/microsoft/hnms.git}.
\end{abstract}

\section{Introduction}
Recent years have seen a great progress on object detection based on deep convolutional neural networks.
The approaches can be roughly categorized as two-stage detectors~\cite{RenHG017,LinDGHHB17} and one-stage detectors~\cite{RedmonF17,LiuAESRFB16,LinGGHD17}. 
In two-stage detectors, a region proposal network (RPN)~\cite{RenHG017} is designed to propose candidate bounding boxes, which are used by the detection head network to refine the bounding box coordinates and to predict the classification scores.
The one-stage detector directly predicts the box coordinates and classification result in one network pass. 
As one of the most essential post-processing steps, non-maximum suppression  (NMS) is used to remove the boxes with similar locations and shapes but lower confidences. 

\begin{figure}[t]
	\centering
	\includegraphics[width=0.85\linewidth]{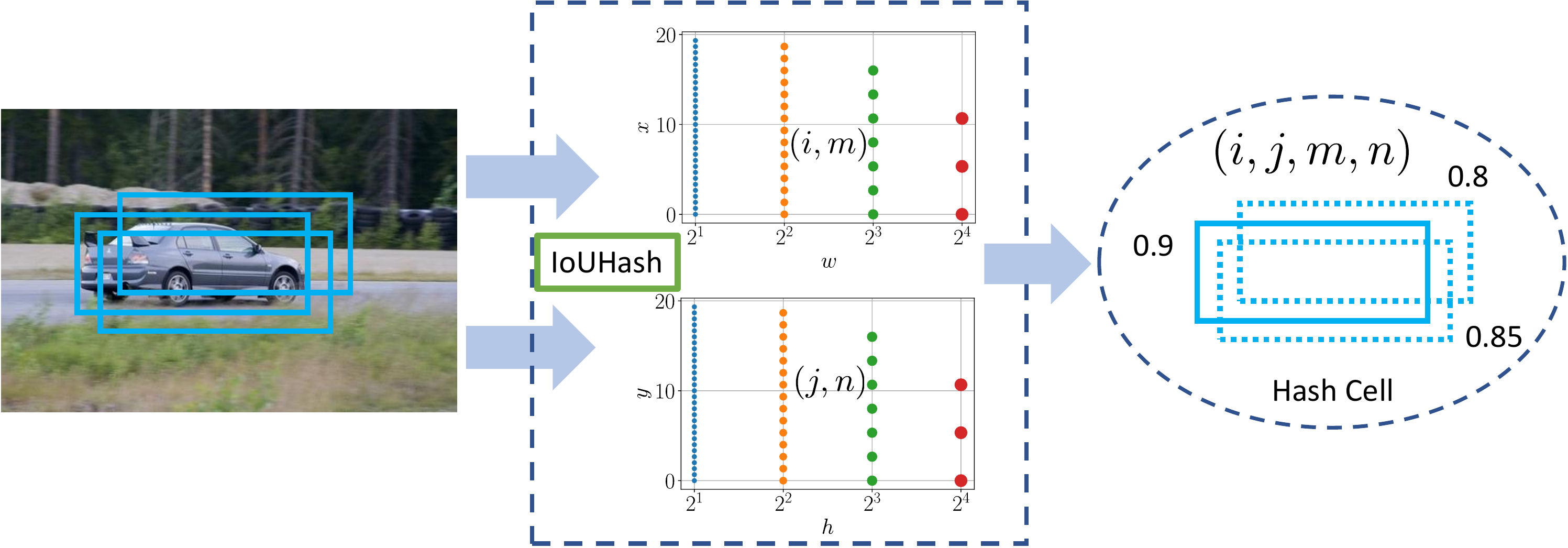}
	\caption{Illustration of Hashing-based NMS process. 
		Each bounding box is hashed
		to four index numbers by the proposed IoUHash function. The box width and height are hashed as $(i, j)\in \mathbb{Z}^2$ based on log scale and the offset is hashed as $(m, n) \in \mathbb{Z}^2$. The dots in the figure means the center of the hash cell.  
		In each hash cell, the box (in solid line) with highest confidence score is kept, and all others (in dotted line) are suppressed. A lower bound can be derived to guarantee the closeness between any two boxes within the same cell.}
	\label{fig:hnms}
\end{figure}

Different variants of NMS have been proposed to improve the detection accuracy
~\cite{BodlaSCD17,sangBS17,LiuLJ15,HeZWS019}.
Instead of discarding the boxes, soft NMS~\cite{BodlaSCD17} decays 
the confidence score as a continuous function of the intersection-over-union (IoU) and keeps all the boxes. 
While the continuous function is manually designed in~\cite{BodlaSCD17}, a special network is learned in~\cite{sangBS17} to rescore the confidence.
Beyond altering the confidence, the bounding box coordinates
are also updated
in~\cite{LiuLJ15,HeZWS019} during suppressing neighboring boxes
to improve the location accuracy. 

For the time cost, the NMS complexity in the worst case is 
$O(N^2)$, where $N$ is the number of boxes. 
Thus, the time cost becomes remarkably high when the number of boxes is large. 
This problem is more severe in object detection with crowded scene as thousands of boxes are generated in RPN for NMS.

To improve inference speed, 
\cite{CaiZWLFAC19} proposed MaxPoolNMS to replace the NMS in RPN by a max pooling operation over the objectness.
It implicitly assumes that boxes from nearby anchor boxes are also similar. 
However, this assumption cannot be guaranteed theoretically because the bounding box regression could change the box coordinates dramatically without any constraint. 

To reduce the time cost and address the issues in existing approaches, we propose a hashing-based NMS (HNMS) approach, which has $O(N)$ time complexity. 
As illustrated in Fig.~\ref{fig:hnms}, the basic idea is to map each box location into a discrete hash value. 
Traditionally, a hash function maps different box locations to different values even if the boxes
are close to each other. Instead, we expect that the boxes at neighboring locations and with similar shapes are mapped to the same hash value, and the boxes located far away should be mapped to 
different hash values.
The region where different boxes are hashed to the same value is denoted as a \textit{hash cell}.
In this way, we can remove the boxes with lower confidence within each cell. 
The idea is similar to the hashing-based nearest neighbor search (NNS) problem~\cite{IndykM98,DatarIIM04,WeissTF08,WangWYL13,WangWSXSL14,WangZSSS18}, 
where similar points are expected to be mapped to similar hashing values. 
In NNS, the metric is normally based on Euclidean distance, and thus the designed function is inappropriate for the detection problem because of the IoU metric. 
For example, two bounding boxes can have small difference in Euclidean distance, but can be far away from each other in IoU.

Considering the definition of IoU, we propose a simple yet effective hashing function, named IoUHash.
The design principle is to make it simple and make the size of each hash cell as similar as possible.  
Given two boxes with the constant IoU, if the size is larger, the offset difference can be farther. 
Thus, we hash the size first (in logarithmic scale) and then the offset (in natural scale) based on the size. 
Analytically, we can derive a lower IoU bound regardless of the cell index if two boxes are within the same hash cell,
which guarantees the suppressed boxes are close enough to the box with highest confidence. 

One issue is that two boxes may be located in adjacent cells and near the boundary. In this case, both boxes may not be suppressed though their actual IoU is large. To address this issue, we apply the HNMS multiple times with different IoUHash functions, where the cell partition parameters are different. 
For two-stage detectors, we directly replace NMS by our HNMS in RPN, and observe no accuracy loss but significant speed gain for the suppression module, e.g. $7.4$x on CARPK in CPU mode. 
For one-stage detector, we apply HNMS as a pre-filtering step which greatly reduces the number of boxes to be filtered by NMS and the overall time cost, e.g. resulting in $6.8$x speed up on CARPK in GPU mode. 

\section{Preliminary}
To make the paper self-contained, we first review the problem that NMS tries to solve and then describe the details of NMS. 

\subsection{Problem Description}
Let $N$ be the number of bounding boxes
and $\mathbf{B} \in \mathbb{R}^{N\times 4}$ be the coordinate matrix,
where each row corresponds to the location of each box. 
Normally, each box can be described by its box size $\{w,h\}$ and center point offset $\{x,y\}$ or its top-left and bottom-right corners $\{x_{left},y_{top},x_{right},y_{bottom}\}$. In this paper, we will use the center offset and size representation $\{w,h,x,y\}$ to describe our algorithm and use \textit{offset} to denote center point offset for simplicity.
A confidence score $s\in [0, 1]$ is attached to every box to indicate the likelihood belonging to the specific class or the general object (e.g. RPN).
Let $\mathbf{s}\in [0,1]^{N}$ be the corresponding confidence vector.
The problem is that multiple boxes with similar shapes may be located for a single object, and we need to keep one and suppress the others.

\renewcommand{\algorithmicrequire}{\textbf{Input:}}
\renewcommand{\algorithmicensure}{\textbf{Output:}}

\subsection{Non-Maximum Suppression}\label{sec:nms}
The alogirthm of NMS can be described as follows. 
All boxes are initialized as \textit{unsuppressed}. 
Then, it goes through each unsuppressed box in a descending order of the confidence score.
IoU is calculated between the current box (with higher confidence) and all the unsuppressed boxes with lower confidences.
The boxes with IoU larger than a pre-defined threshold are suppressed. 
This process will continue until all boxes are checked.  

The sorting takes $O(N\log(N))$ time complexity.
If each unsuppressed box can suppress $S$ boxes on average, the time complexity of the two loops is $O(N^2/(S + 1))$.
In the worst case where no box is suppressed, the complexity is $O(N^2)$. 
If all the other boxes are suppressed by the first box (highest confidence score), the time complexity is $O(N)$.
Thus, the time cost ranges from $O(N\log(N)) + O(N)$
to $O(N\log(N)) + O(N^2)$. 
Next, we will present our proposed approach which has $O(N)$ time complexity independent of the data distribution. 

\section{Proposed Approach}
The core idea is to quantize the continuous-valued box coordinates to 
discrete values and then perform non-maximum suppression within each 
hash cell. Thus, we first introduce the hashing function 
in Sec.~\ref{sec:iouhash} and then 
the suppression logic in Sec.~\ref{sec:hnms}.
Sec.~\ref{sec:discussion} gives a discussion on the performance. 

\subsection{IoUHash Function} \label{sec:iouhash}
To make it simple, we design the function to be data-independent, i.e. no parameters are required to learn from the data.
Another principle is to make the expected IoU roughly the same 
if any two boxes are hashed to the same cell.
In this way, all hash cells can be treated equally.  

Based on the IoU definition, if the widths and heights of two boxes are both larger, their offsets can be farther to have the same IoU. 
If two boxes are both small, a slight offset change would lead to large IoU change. 
To make the IoU the same, the offset quantization should be based on the box size. 
Thus, we propose to quantize the size first and then the offset.  

\begin{figure}[t!]
	\centering
	\begin{tabular}{c@{~~~~~~~~~}c}
	\includegraphics[width=0.4\linewidth]{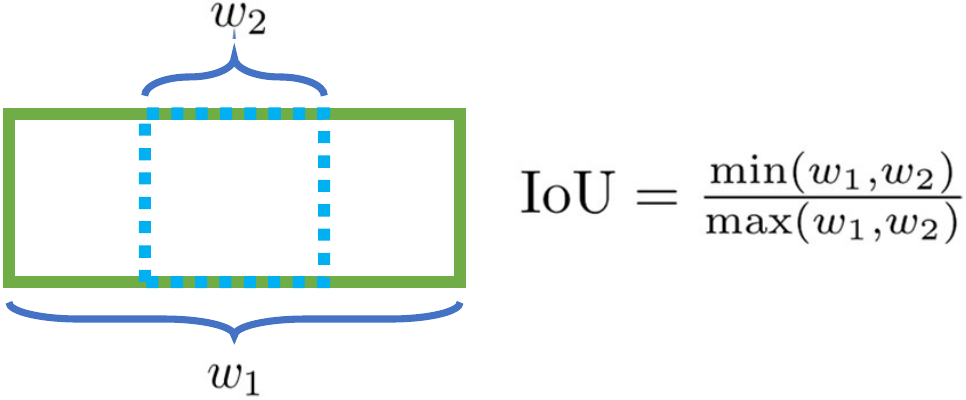}	&
	\includegraphics[width=0.4\linewidth]{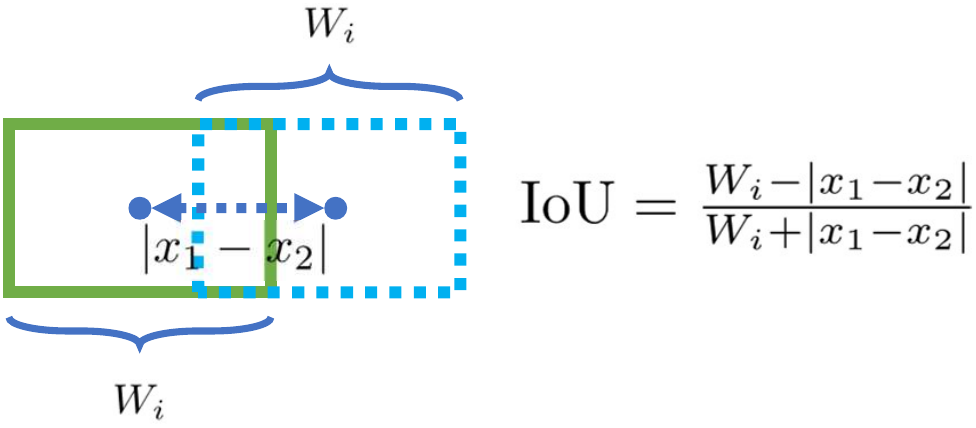} \\
	\scriptsize{(a)} & \scriptsize{(b)}
	\end{tabular}
\caption{IoU calculation for deriving IoUHash.}
\label{fig:iou}
\end{figure}

The width and height are quantized independently for simplicity. 
Take the width as an example. 
If two boxes are of the same height and offsets, their IoU can be written as 
$\min(w_{1}, w_{2}) / \max(w_{1}, w_{2})$, where $w_{1}$ and $w_{2}$ are the widths of the boxes, as illustrated in Fig. \ref{fig:iou} (a).
This motivates us to design the function such that 1) the width is partitioned into multiple
disjoint ranges, and 2) the ratio of the range centers should be the same if we attach a center point to each range. 
Let $\{W_{i}, i\in \mathbb{Z}\}$ be the range centers. 
That is, we should have
$\alpha \triangleq W_{i} / W_{i + 1}, \forall i$, where $\alpha$ is the pre-defined parameter, or
\begin{align}
	W_{i} = {W_{0}}/{\alpha^{i}}, i \in \mathbb{Z},
	\label{eqn:wi}
\end{align}
where $W_0$ is the $0$-th cell center as a parameter.
Note that the index of $i$ can be smaller than $0$. 
With a log operation, we have
$\log(W_{i}) = \log(W_{0}) - i\log(\alpha_{w})$.
In other words, the width is equally partitioned in the log-scale.
Given any box with width $w$, we quantize the width as the $i$-th cell if 
\begin{align}
	i = \lfloor{(\log(W_0) - \log(w))}/{\log(\alpha)}\rceil,
	\label{eqn:i}
\end{align}
where $\lfloor\rceil$ means the integer round operation. 
With this design, if two boxes are of the same height and offset, and with widths being two adjacent range centers (e.g. $W_i$ and $W_{i + 1}$), their IoU is always $\alpha$ for any $i$. 
If their widths also falls into the same range (e.g. $W_i$), 
the minimum IoU is also $\alpha$. 
Thus, $\alpha$ represents the distance between adjacent cells and the cell size. 
Similarly, the height $h$ is quantized to the $j$-th cell if
\begin{align}
	j & = \lfloor{(\log(H_0) - \log(h))}/{\log(\alpha)}\rceil,
	\label{eqn:j}
\end{align}
where $H_0$ is the $0$-th cell for the height.

After quantizing the width and height, we also hash the 
$x$-offset and $y$-offset independently. 
Take the $x$-offset for instance. 
Given two boxes, assume that the widths are identical and equal to $W_i$ (Eqn.~\ref{eqn:wi}),
the heights and $y$-offsets are also the same, but $x$-offsets ($x_1$ and $x_2$) are different. 
Then, the IoU is $(W_i - |x_1 - x_2|)/(W_i  + |x_1 - x_2|)$, shown in Fig. \ref{fig:iou} (b) if $|x_1 - x_2| \le W_i$, 
and $0$, otherwise.
As we can see, the IoU is only related to the distance between the centers in the $x$ direction in this case.
This leads us to quantize the $x$-axis equally. Let $X_m$ be the partition center, 
and $\delta_i \triangleq X_{m + 1} - X_{m}$ 
be the distance between any two adjacent partition centers. 
Note, the $\delta_i$ depends on $W_i$, which represents how wide the box is. 
If two boxes falls into two adjacent partition centers, their IoU is
designed to be $\alpha$ (recall that $\alpha$ represents the distance of the adjacent width/height cells), i.e. $\alpha = (W_i - \delta_i)/(W_i + \delta_i)$.
Then, we have 
\begin{align}
	\delta_i & = W_i{(1 - \alpha)}/{(1 + \alpha)}
	\label{eqn:delta_i} \\
	X_m & = b_x\delta_i + m \delta_i, m \in \mathbb{Z}
\end{align}
where $b_x$ is a parameter. 
Given the horizontal offset as $x$, we hash it to 
\begin{align}
	m = \lfloor{x}/{\delta_i} - b_x\rceil.
	\label{eqn:m}
\end{align}


Similarly, the vertical offset $y$ is quantized as
\begin{align}
	n = \lfloor{y}/{\delta_j} - b_y\rceil,
	\label{eqn:n}
\end{align}
where
\begin{align}
	\delta_j &= H_j(1 - \alpha)/(1 + \alpha), 
	\label{eqn:delta_j} \\
	H_j &= H_0/\alpha^j.
	\label{eqn:hj}
\end{align}

The algorithm is described in Alg.~\ref{alg:hash}.
In summary, given any two boxes, if any three dimensions (e.g. width, height, $x$-offset) are the same and equal to the corresponding cell center, 
and the other dimension are equal to the adjacent cell centers, 
the IoU is alway $\alpha$. If the unequal dimension is also hashed to the
same cell, the minimum IoU is also always $\alpha$.

\begin{algorithm}[t]
	\caption{IoUHash}
	\label{alg:hash}
	\begin{algorithmic}[1]
		\REQUIRE
		Bounding box $(w, h, x, y)$, and hyper-parameters 
		$W_0$, $H_0$, $b_x$, $b_y$, $\alpha$
		\ENSURE
		Hash code ($i, j, m, n$)
		\STATE{Calculate $i$ based on Eqn.~\ref{eqn:i} with $W_0$ and $\alpha$}
		\STATE{Calculate $j$ based on Eqn.~\ref{eqn:j} with $H_0$ and $\alpha$}
		\STATE{Calculate $W_{i}$ based on Eqn~\ref{eqn:wi} with $W_0$ and $i$}
		\STATE{Calculate $H_{j}$ based on Eqn~\ref{eqn:hj} with $H_0$ and $j$}
		\STATE{Calculate $\delta_{i}$ based on Eqn~\ref{eqn:delta_i} with $W_i$ and $\alpha$}
		\STATE{Calculate $\delta_{j}$ based on Eqn~\ref{eqn:delta_j} with $H_j$ and $\alpha$}
		\STATE{Calculate $m$ based on Eqn.~\ref{eqn:m} with $x, \delta_i$ and $b_x$}
		\STATE{Calculate $n$ based on Eqn.~\ref{eqn:n} with $y, \delta_j$ and $b_y$}
		\STATE{return ($i, j, m, n$)}
	\end{algorithmic}
\end{algorithm}

\subsubsection{Relation with Related Work}
This idea of quantizing the box is similar to the anchor size design 
in the detection framework, e.g. in FPN~\cite{LinDGHHB17}, Faster R-CNN \cite{RenHG017}. 
Each ground truth box is assigned to different anchors during training, which is analogous to the process of hashing each box to different cells. 
One major difference is that the anchor sizes are designed jointly. 
For example in Faster R-CNN\cite{RenHG017}, the aspect ratio of width and height is set to be $1:1$, $2:1$ or $1:2$. 
Comparably, we design the width and height independently. 
If each component has $Q$ different cells, we can have as many as $Q^2$
different \textit{anchors}, which is normally much larger than the number of anchor shapes. 

MaxPoolNMS~\cite{CaiZWLFAC19} performs the suppression by a max pooling over the objectness in RPN. 
If we treat anchors as hash cells,
the approach can be interpreted as hashing each proposal to its corresponding anchor and suppressing other boxes in adjacent cells. 
The quantization here implicitly ignores the bounding box regression, which can change the box location without any constraint. 
Comparably, we perform hashing on regressed boxes to make each cell more compact.  
Another difference is that we suppress boxes within the same cell rather than in adjacent cells. 

\subsubsection{IoU Upper and Lower Bound}\label{sec:bound}
If two bounding boxes are mapped to the same cell,
the upper IoU bound is $1$ if the two boxes are identical.
Next, we calculate the lower bound. 

Let $(w_1, h_1, x_1, y_1)$ and $(w_2, h_2, x_2, y_2)$ be two boxes, which are quantized to the same cell. Then, the intersection area can be written as
$I = F(r - l)F(b - t)$, where 
$F(x) = x$ if $x > 0$ and $0$, otherwise; and 
\begin{align}
	l & = \max(x_1 - \frac{1}{2}w_1, x_2 - \frac{1}{2}w_2),
	r = \min(x_1 + \frac{1}{2}w_1, x_2 + \frac{1}{2}w_2) \\
	t & = \max(y_1 - \frac{1}{2}h_1, y_2 - \frac{1}{2}h_2),
	\textbf{ } b = \min(y_1 + \frac{1}{2}h_1, y_2 + \frac{1}{2}h_2).
\end{align}
If their centers are far enough, their intersection can vanish to $0$ and the smallest IoU is $0$. 
To avoid such cases, we should have the following condition (omitting the requirement on $y_1$, $y_2$) hold always:
\begin{align}
	x_1 + \frac{1}{2}w_1 > x_2 - \frac{1}{2}w_2,
	x_2 + \frac{1}{2}w_2 > x_1 - \frac{1}{2}w_1,
\end{align}
which is equivalent to 
\begin{align}
	|x_1 - x_2| < \frac{1}{2}(w_1 + w_2).
\end{align}
Since the two boxes are in the same cell, the largest value of $|x_1 - x_2|$ is equal to $\delta_i$ if they fall into the $i$-th width cell 
according to Eqn. \ref{eqn:i}. 
The smallest value of $w_1$ (or $w_2$) is $\alpha^{0.5} W_i$, which is the boundary between $W_{i - 1}$ and $W_i$.
Thus, if we have the following condition hold
\begin{align}
	(1 - \alpha)/(1 + \alpha) < \alpha^{0.5},
	\label{eqn:nonzero_condition}
\end{align}
their IoU is always larger than $0$. 
With $\alpha = 0.3$, Eqn.~\ref{eqn:nonzero_condition} holds. 
When $\alpha$ is increased, the right side is larger but the left side is smaller, which means the requirement is always satisfied when $\alpha \geq 0.3$.
Next, we derive the lower bound under this condition. 

\begin{wrapfigure}{R}{.3\textwidth}
	\includegraphics[width=.3\textwidth]{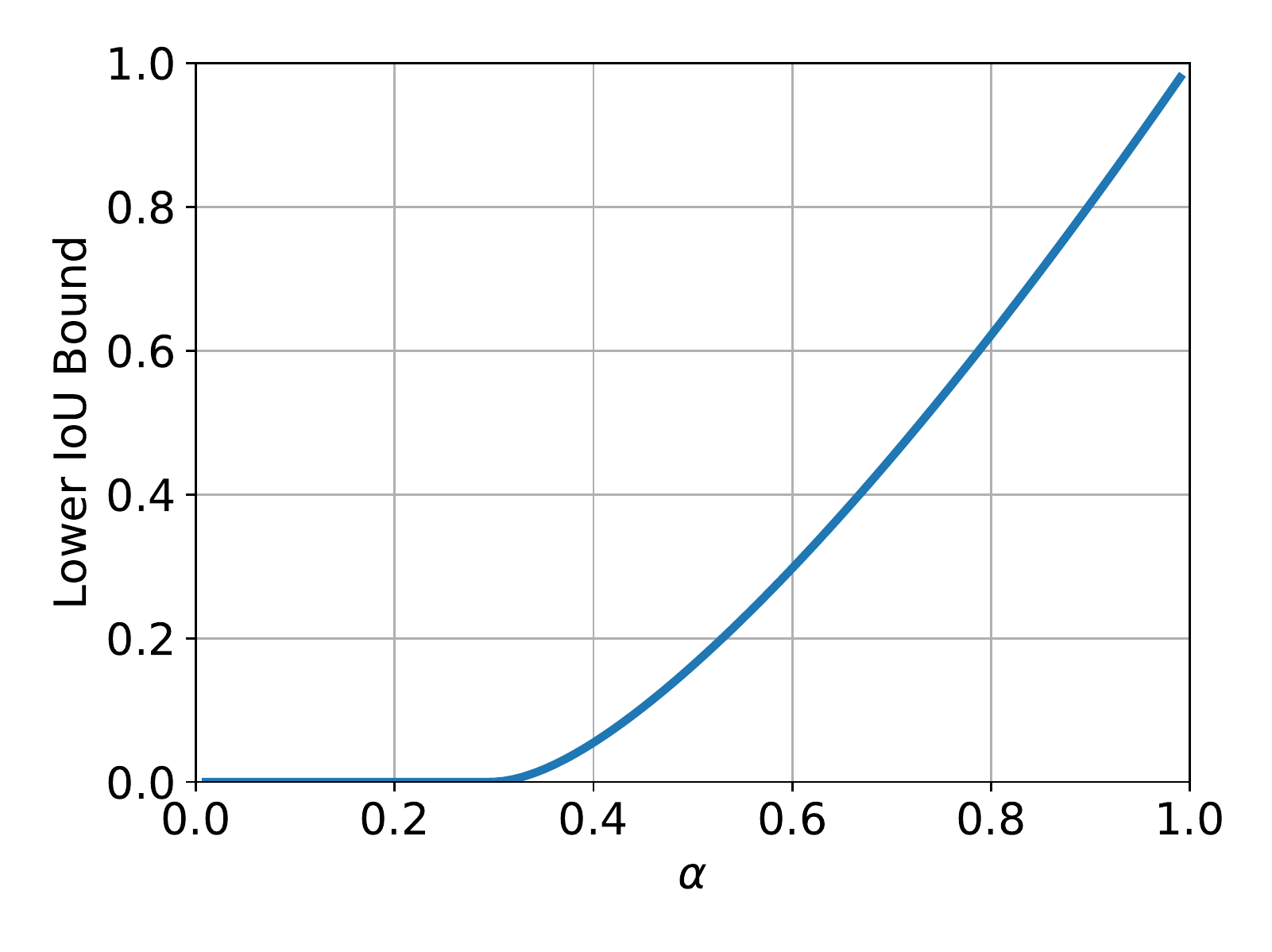}
	\caption{Lower IoU bound as a function of $\alpha$.}
	\label{fig:low}
\end{wrapfigure}

Due to the same cell the two boxes are quantized into, we can express each component by the distance to the cell center as 
\begin{align}
	\begin{split}
		\label{eqn:trick}
		w_k & = {W_0}/{\alpha^{i + i_k}}, \text{~~~~~~~~~~~~} i_k \in [-0.5, 0.5) \\
		h_k & = {H_0}/{\alpha^{j + j_k}}, \text{~~~~~~~~~~~~} j_k \in [-0.5, 0.5) \\
		x_k & = b_x\delta_i + (m + m_k)\delta_i,  m_k \in [-0.5, 0.5) \\
		y_k & = b_y\delta_j + (n + n_k)\delta_j, \text{~~} n_k \in [-0.5, 0.5)
	\end{split}
\end{align}
where $k = 1, 2$, $i, j, m, n$ is the hash code for each dimension. 
By substituting Eqn.~\ref{eqn:trick} to the IoU definition,
we can easily conclude that
the IoU has no relation with $b_x, b_y, W_0, H_0, i, j, m, n$, 
but only depends on 
$\alpha$, and $i_k$, $j_k$, $m_k$ and $n_k$, which are all bounded from -$0.5$ to $0.5$. 
Meanwhile, we can see that the minimum IoU must reside in one of the boundaries, where $i_k$, $j_k$, $m_k$, and $n_k$ equals -$0.5$ or $0.5$. 
Instead of sticking to a closed form of lower bound, we calculate the lower bound by 1) selecting any $b_x, b_y, W_0, H_0, i, j, m, n$ since the IoU is 
constant with these values, 2) enumerating all the combinations of different 
$i_k$, $j_k$, $m_k$ and $n_k$ (equal -$0.5$ or $0.5$), which means $2^8$
different combinations for two boxes, 3) calculating the bounding box coordinates by Eqn.~\ref{eqn:trick} for each combination; 4) calculating the IoU and choosing the minimum IoU, which is the lower bound.
The algorithm flow and more details can be found in Appendix \ref{sec:iou_lower_bound}.

Fig.~\ref{fig:low} shows the lower IoU bound with different $\alpha$.
As we can see, the bound is a monotone non-decreasing function
of $\alpha$. When $\alpha = 0.3$, the bound is $1.4e^{-4}$,
and thus if $\alpha \ge 0.3$, the IoU between any two boxes within 
the same hash cell is guaranteed to be larger than $0$.
This conclusion is consistent with satisfying Eqn.~\ref{eqn:nonzero_condition}.



\subsection{Hashing-based Non-Maximum Suppression} \label{sec:hnms}
After mapping each box to a hash cell by IoUHash, we simply keep the box with the largest confidence score and remove all others within each cell. 
The process can be described in Alg.~\ref{alg:hnms}. 

\begin{algorithm}[t]
	\caption{Hashing-based Non-Maximum Suppression}
	\label{alg:hnms}
	\begin{algorithmic}[1]
		\REQUIRE
		Bounding box $\mathbf{B} \in \mathbb{R}^{N\times 4}$,
		confidence score $\mathbf{s}\in [0.,1.]^{N}$,
		$\alpha$, $b_x, b_y, W_0, H_0$
		\ENSURE
		Indices of kept boxes
		\STATE{$\mathbf{C}$ = IoUHash($\mathbf{B}$) (Alg.~\ref{alg:hash}) with 
			$\alpha$, $W_0, H_0, b_x, b_y$}
		\STATE{code2idx = \{\}}
		\FOR{$i$ in $[0, N - 1]$}
		\IF{$\mathbf{C}_i$ in code2idx}
		\STATE{pre = code2idx[$\mathbf{C}_i$]}
		\IF{$s[pre] < s[i]$}
		\STATE{$\text{code2idx}[\mathbf{C}_i] = i$}
		\ENDIF
		\ELSE
		\STATE{$\text{code2idx}[\mathbf{C_i}] = i$}
		\ENDIF
		\ENDFOR
		\STATE{return code2idx.values()}
	\end{algorithmic}
\end{algorithm}


Though we have a lower bound to guarantee the lowest IoU between any two boxes within the same cell, it cannot guarantee any two boxes with IoU larger than
the lower bound falls to the same cell. 
For example, two boxes are close enough and have high IoU, but they can be hashed into two adjacent hash cells though they are very close to the boundary of the two adjacent cells.
To solve the problem, we apply HNMS processes multiple times with different parameters. 
In IoUHash, $\alpha$ controls the size of each cell, while
$W_0$, $H_0$, $b_x, b_y$ controls the offset. 
Given $\alpha$, 
we equally split the space to generate multiple IoUHash functions. 
Let $K$ be the number of IoUHash functions. The $k$-th ($k\in [0, K-1]$) IoUHash's parameters ($W_0^{(k)}, H_0^{(k)}, b_x^{(k)}, b_y^{(k)}$) is calculated by satisfying the following condition, 
\begin{align}
	\begin{split}
		\log(W_0^{(k)}) & = -\log(\alpha) {k}/{K}, 
		\log(H_0^{(k)}) = -\log(\alpha) {k}/{K} \\
		b_x^{(k)} & = {k}/{K}, \text{~~~~~~~~~~~~~~~~~~} b_y^{(k)} = {k}/{K}.
	\end{split}
	\label{eqn:i_hash}
\end{align}
That is, $\log(W_0^{(k)})$ equally splits the range from
$\log(1)$ to $\log(1) + (-\log(\alpha))$, and $b_x^{(k)}$
equally splits the range from 0 to 1. 
The whole process with multi HNMS is illustrated in Alg.~\ref{alg:multi_hnms}.

\begin{algorithm}[t]
	\caption{Multi HNMS}
	\label{alg:multi_hnms}
	\begin{algorithmic}[1]
		\REQUIRE
		Bounding box $\mathbf{B} \in \mathbb{R}^{N\times 4}$,
		confidence score $\mathbf{s}\in [0.,1.]^{N}$, $K$, $\alpha$.
		\ENSURE
		Indices of kept boxes
		\STATE{Init an empty array of HNMS with length $K$}
		\FOR{$k$ in $[0, K - 1]$}
		\STATE{Calculate ($W_0^{(i)}, H_0^{(i)}, b_x^{(i)}, b_y^{(i)}$) based on Eqn.~\ref{eqn:i_hash}}
		\STATE{HNMS[i] = IoUHash with the parameter of $W_0^{(i)}$}
		\ENDFOR
		\STATE{keep=[1: N - 1]}
		\FOR{$i$ in $[0, K - 1]$}
		\STATE{currkeep = HNMS[i]($\mathbf{B}$, $\mathbf{s}$) based on Alg.~\ref{alg:hnms}}
		\STATE{keep = keep[currkeep]; $\mathbf{B} = \mathbf{B}[\text{currkeep]}$; $\mathbf{s} = \mathbf{s}[\text{currkeep]}$}
		\ENDFOR
		\STATE{return keep}
	\end{algorithmic}
\end{algorithm}

\subsection{Discussion}\label{sec:discussion}
\subsubsection{Approximation.}
Obviously, the complexity of HNMS is $O(N)$, which is faster than
the vanilla NMS ($O(N\log(N) + O(N^2 / (S + 1)))$).
If all boxes are split into positive boxes and negative boxes based on 
the NMS filter result,  
HNMS can be regarded as an approximate process of NMS,
and has the following misaligned cases. 

First, if the lower bound is lower than the NMS threshold, within each cell, HNMS could suppress the boxes with lower IoU, which leads to lower recalls. For this problem, we need a smaller hash cell or a higher $\alpha$.
Second, if the lower bound is higher than threshold, HNMS may fail to suppress the boxes whose IoU is larger than the threshold. This can lead to false positives or lower precision. 
For this problem, we need a larger hash cell or lower $\alpha$. 
Third, if the lower bound is exactly the same as the threshold, HNMS might still fail to suppress some negative boxes or fail to keep some positive boxes. 
The reason is that NMS always suppresses boxes by unsuppressed boxes, while HNMS suppresses boxes within each cell and the box with the highest confidence might be a suppressed box in NMS. 
For example, we have three boxes 
A (100, 100, 54.1, 50), 
B (100, 100, 79.1, 50), 
C (100, 100, 96.1, 50). 
The four numbers are width, height, center $x$-offset and center $y$-offset. 
The confidence scores are $0.9$, $0.8$, $0.7$, respectively. 
For HNMS, $W_0 = H_0 = 1$, $b_x=b_y= 0$, 
$\alpha = 0.73$. The lower IoU bound is $0.5015$, 
which is set as the NMS threshold. The IoU between A and B is $0.7100$, and thus B is suppressed. 
The IoU between A and C is $0.4934$, and is lower than the threshold. Thus, the result is A and C. 
Based on Alg.~\ref{alg:hnms}, the result is A and B because B and C are in the same cell. 
Thus, C is missed and B is kept. 

Since HNMS is not exactly equivalent to NMS, we apply HNMS in the following way. In two-stage detectors, we directly replace NMS in RPN by HNMS and observe significant speed up without any accuracy loss. 
The reason is that 
the RoI head is able to fix imperfect proposals. 
NMS in the RoI head is not changed since we find the time cost of NMS
is minor. 
In one-stage detectors, we insert the HNMS as a pre-filter before applying the NMS, which greatly reduces the number of boxes for NMS and the overall (HNMS + NMS) time cost can also be reduced significantly. 

\subsubsection{Implementation.} Based on Sec.~\ref{sec:iouhash} and Sec.~\ref{sec:hnms}, we can easily implement the CPU code. For GPU, the challenging part is 
the suppression logic since different hash cells can have different numbers of boxes. For this problem, we implement it in the following way. First, we convert each hash code ($i, j, m, n$)
to a unique integer. 
Second, we find the unique hash code and the 
index of the unique hash code based on the integer. 
Third, we calculate the maximum confidence score 
for each unique code by the atomic max operation.
Finally, we find the index of the box with maximum score
by the atomic compare-and-swap operation.
Further details can be found in Appendix \ref{sec:gpu_impl}.

\section{Experiment}

\begin{table}[t!]
\centering
\caption{Experimental results with Faster-RCNN-R50-FPN. RPN-NMS and Total represent the time cost in ms. Speed is relative to RPN-NMS.}
\label{tbl:rpn}
\scalebox{\scaletable}{
\begin{tabular}{c}
	(a) CARPK \\
\begin{tabular}{c@{~~}c@{~~}c@{~~~}ccc@{~~~}ccc}
\toprule
 &  &  & \multicolumn{3}{c}{CPU} & \multicolumn{3}{c}{GPU}\\
\cmidrule(lr){4-6}
\cmidrule(lr){7-9}
 & $K$ & mAP & RPN-NMS & Speed & Total & RPN-NMS & Speed & Total\\
\midrule
NMS &  & 96.9 & $223.3\pm 36.5$ & 1x & 3935.7 & $27.2\pm 2.0$ & 1x & 149.0\\
\midrule
\multirow{4}{*}{HNMS} & 1 & \textbf{97.1} & $30.3\pm 5.4$ & \textbf{7.4x} & 3665.6 & $6.5\pm 1.0$ & \textbf{4.2x} & 132.8\\
 & 2 & \textbf{97.1} & $49.0\pm 6.5$ & \textbf{4.6x} & 3571.5 & $9.3\pm 1.2$ & \textbf{2.9x} & 137.9\\
 & 3 & 97.0 & $52.3\pm 7.6$ & 4.3x & 3790.5 & $11.2\pm 1.5$ & 2.4x & 136.9\\
 & 4 & 96.9 & $68.8\pm 9.6$ & 3.2x & 3790.6 & $13.2\pm 1.7$ & 2.1x & 136.2\\
\bottomrule
\end{tabular}
\\
\\
(b) SKU-110K
\\
\begin{tabular}{c@{~~}c@{~~}c@{~~~}ccc@{~~~}ccc}
\toprule
 &  &  & \multicolumn{3}{c}{CPU} & \multicolumn{3}{c}{GPU}\\
\cmidrule(lr){4-6}
\cmidrule(lr){7-9}
 & $K$ & mAP & RPN-NMS & Speed & Total & RPN-NMS & Speed & Total\\
\midrule
NMS &  & \textbf{90.5} & $478.4\pm 64.4$ & \textbf{1x} & 2599.9 & $32.2\pm 3.7$ & \textbf{1x} & 133.7\\
\midrule
\multirow{4}{*}{HNMS} & 1 & 89.8 & $41.4\pm 5.8$ & 11.6x & 2191.5 & $5.9\pm 0.6$ & 5.5x & 101.4\\
 & 2 & 90.4 & $69.2\pm 7.4$ & 6.9x & 2220.2 & $8.6\pm 0.6$ & 3.7x & 103.8\\
 & 3 & \textbf{90.5} & $79.9\pm 11.8$ & \textbf{6.0x} & 2250.3 & $11.4\pm 1.3$ & \textbf{2.8x} & 107.8\\
 & 4 & \textbf{90.5} & $102.6\pm 11.8$ & \textbf{4.7x} & 2239.9 & $17.1\pm 1.9$ & \textbf{1.9x} & 101.4\\
\bottomrule
\end{tabular}
\\
\\
(c) CrowdHuman
\\
\begin{tabular}{c@{~~}c@{~~}c@{~~~}ccc@{~~~}ccc}
\toprule
 &  &  & \multicolumn{3}{c}{CPU} & \multicolumn{3}{c}{GPU}\\
\cmidrule(lr){4-6}
\cmidrule(lr){7-9}
 & $K$ & mAP & RPN-NMS & Speed & Total & RPN-NMS & Speed & Total\\
\midrule
NMS &  & \textbf{82.1} & $307.9\pm 48.4$ & \textbf{1x} & 3643.5 & $27.3\pm 4.3$ & \textbf{1x} & 148.8\\
\midrule
\multirow{4}{*}{HNMS} & 1 & 81.8 & $38.9\pm 6.6$ & 7.9x & 3379.8 & $6.3\pm 0.8$ & 4.3x & 101.8\\
 & 2 & 82.0 & $53.8\pm 6.7$ & 5.7x & 3584.5 & $10.4\pm 1.3$ & 2.6x & 114.4\\
 & 3 & 82.0 & $64.1\pm 9.4$ & 4.8x & 3530.9 & $12.6\pm 0.8$ & 2.2x & 110.6\\
 & 4 & \textbf{82.1} & $73.0\pm 10.9$ & \textbf{4.2x} & 3434.4 & $15.4\pm 0.9$ & \textbf{1.8x} & 113.7\\
\bottomrule
\end{tabular}
\\
\end{tabular}
}
\end{table}

\begin{figure}
	\centering
	\begin{tabular}{@{}c@{}c@{}c@{}}
		\includegraphics[width=\thirdfigurelength\linewidth]{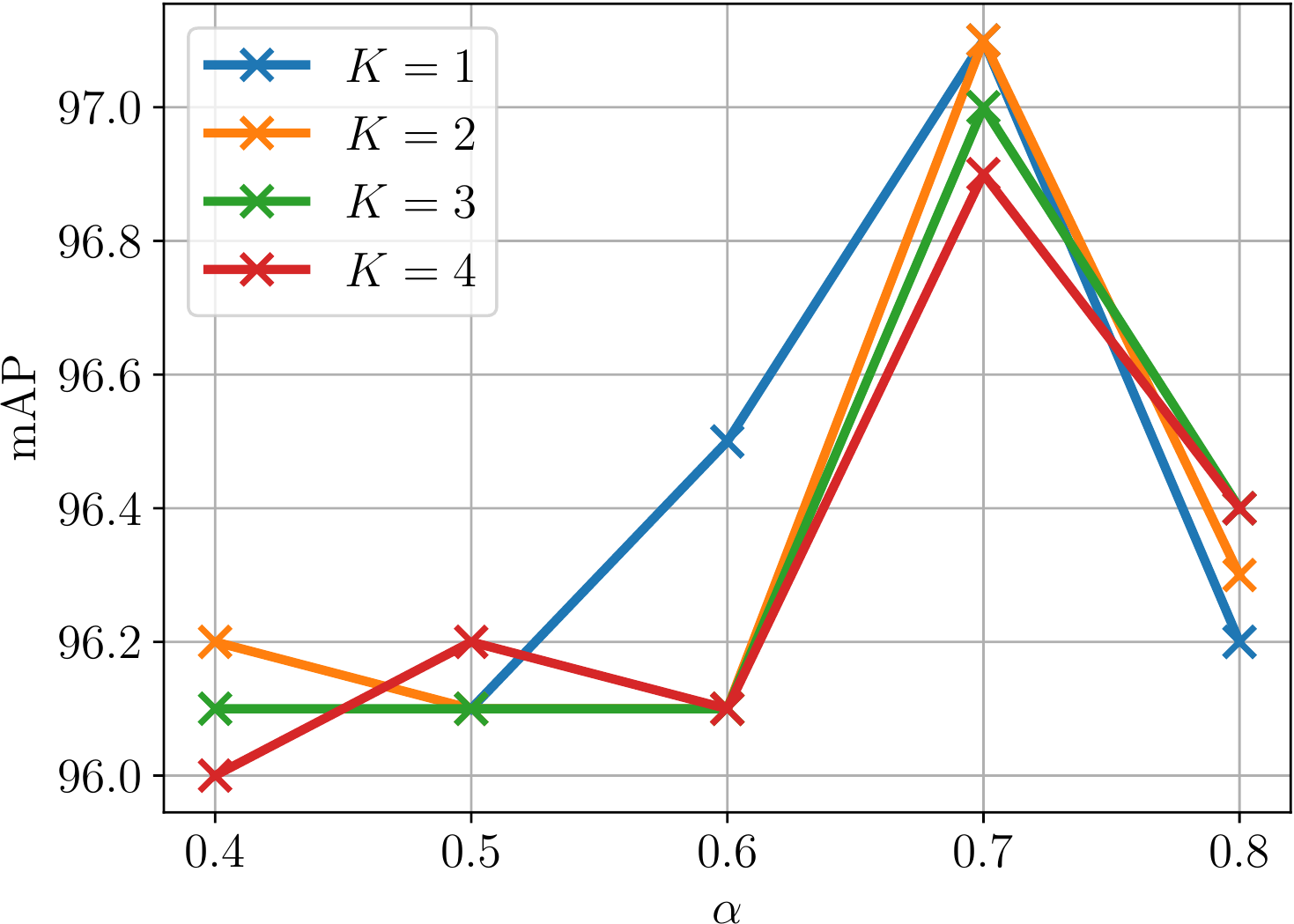}
		&
		\includegraphics[width=\thirdfigurelength\linewidth]{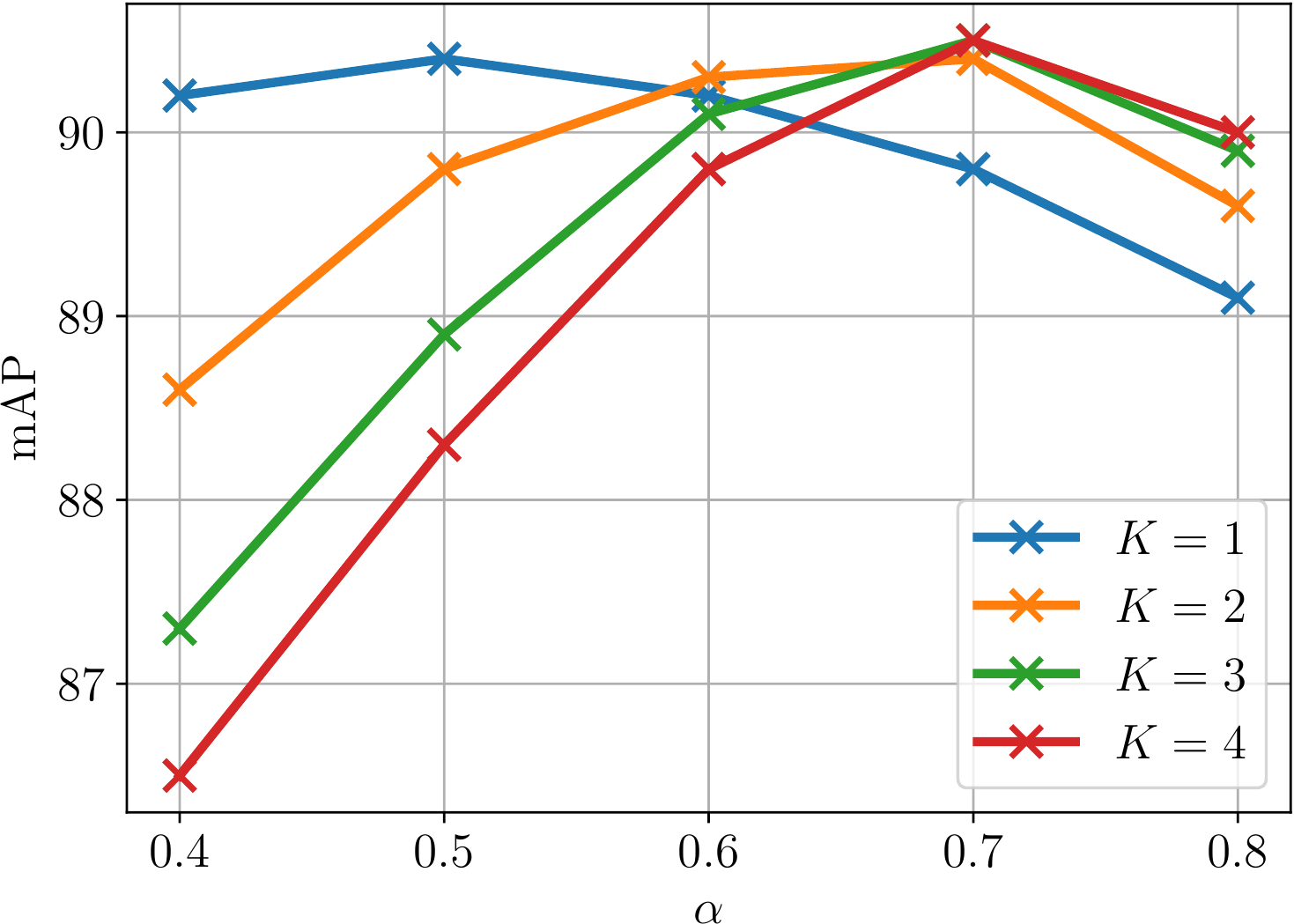}
		&
		\includegraphics[width=\thirdfigurelength\linewidth]{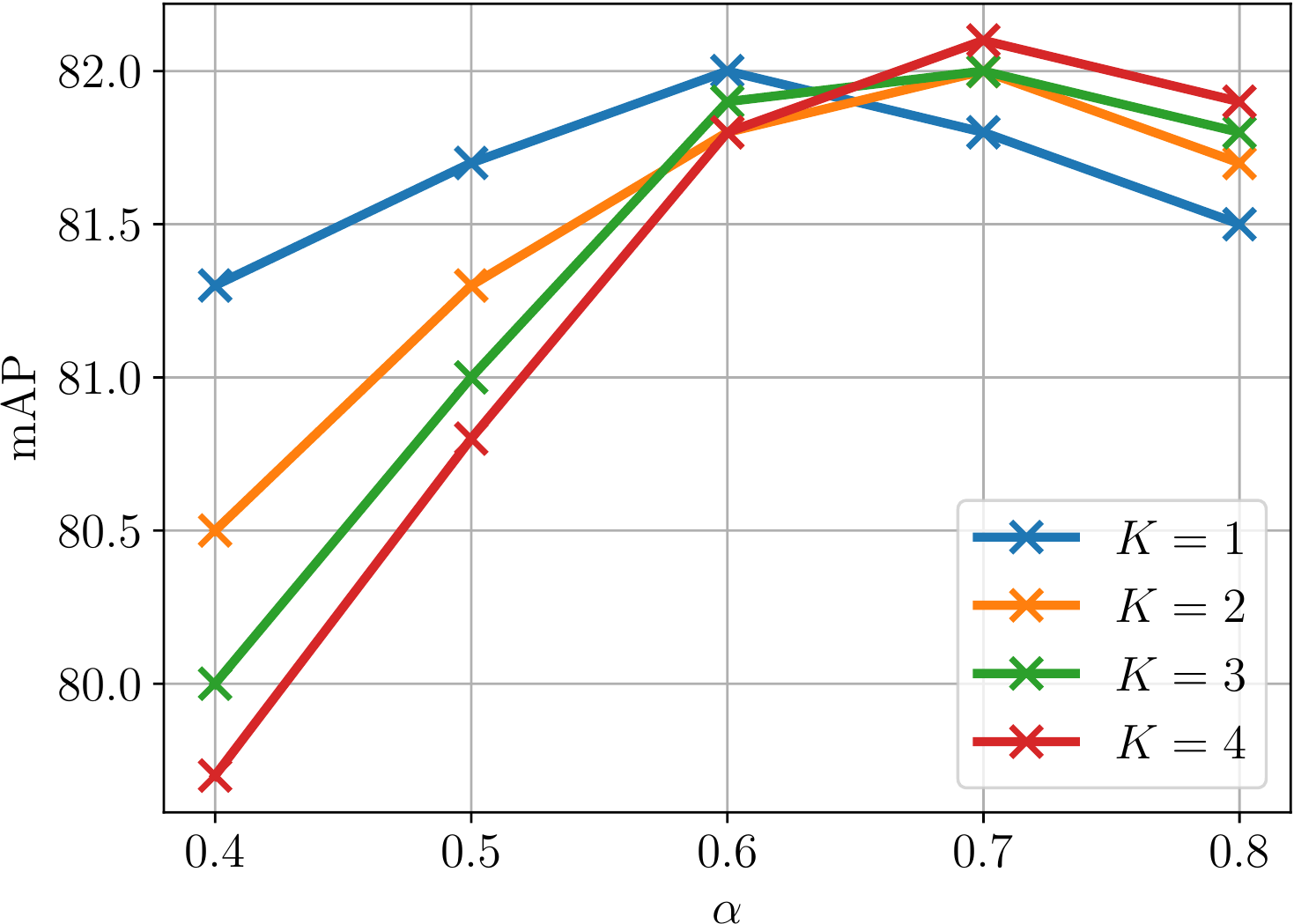} \\
		\scriptsize{(a) CARPK}
		&
		\scriptsize{(b) SKU-110K}
		&
		\scriptsize{(c) CrowdHuman}
	\end{tabular}
	\caption{Accuracy with different $\alpha$ and $K$ for HNMS with Faster-RCNN-R50-FPN.}
	\label{fig:acc_alpha}
\end{figure}

\subsection{Settings}
We conduct experiments on three datasets: CARPK~\cite{HsiehLH17}, 
SKU-110K~\cite{GoldmanHEGH19}, and CrowdHuman~\cite{abs-1805-00123}.
CARPK is a car parking lot dataset, which contains $989$ training images 
with $43$ boxes/image
and $459$ test images with 
$103$ boxes/image.
SKU-110K was collected in retail environment for product item detection,
which provides $8.2$K training images with 147 boxes per image and $2,941$ testing images with 
147 boxes/image. 
CrowdHuman is a benchmark dataset for crowded person detection, which 
has
$15$K training images with 29 boxes/image and $4,730$ validation images with 27 boxes/image.
It provides visible person box, full body box and human head annotations. Here, we use the visible person box annotation. The region marked as \textit{mask} is ignored in evaluation and is removed during training.  
Pascal VOC~\cite{everingham2015pascal} and COCO~\cite{lin2014microsoft} are two common datasets for general object
detection, where the number of boxes per image is around
3 boxes/image and 7 boxes/image, respectively. 
We do not show the results on these two datasets
here
but in Appendix \ref{sec:voc_coco}
because the number of boxes is not large, and the time cost of NMS is minor. 


All the models are trained on the training set and evaluated on the test set or the validation set. 
Mean average precision (mAP) at IoU threshold $0.5$ is used for accuracy comparison. 
The speed is evaluated on a workstation with Intel(R) Xeon(R) CPU E5-2620 v4 @ 2.10GHz and TITAN XP.
Both the NMS and the HNMS are implemented in C++/Cuda. 
Pytorch and Maskrcnn-Benchmark are used as the deep learning toolkit. 
The time cost is calculated based on the first $100$ images. 


\begin{figure}[t!]
	\centering
	\begin{tabular}{cc}
		\includegraphics[width=\halffigurelength\linewidth]{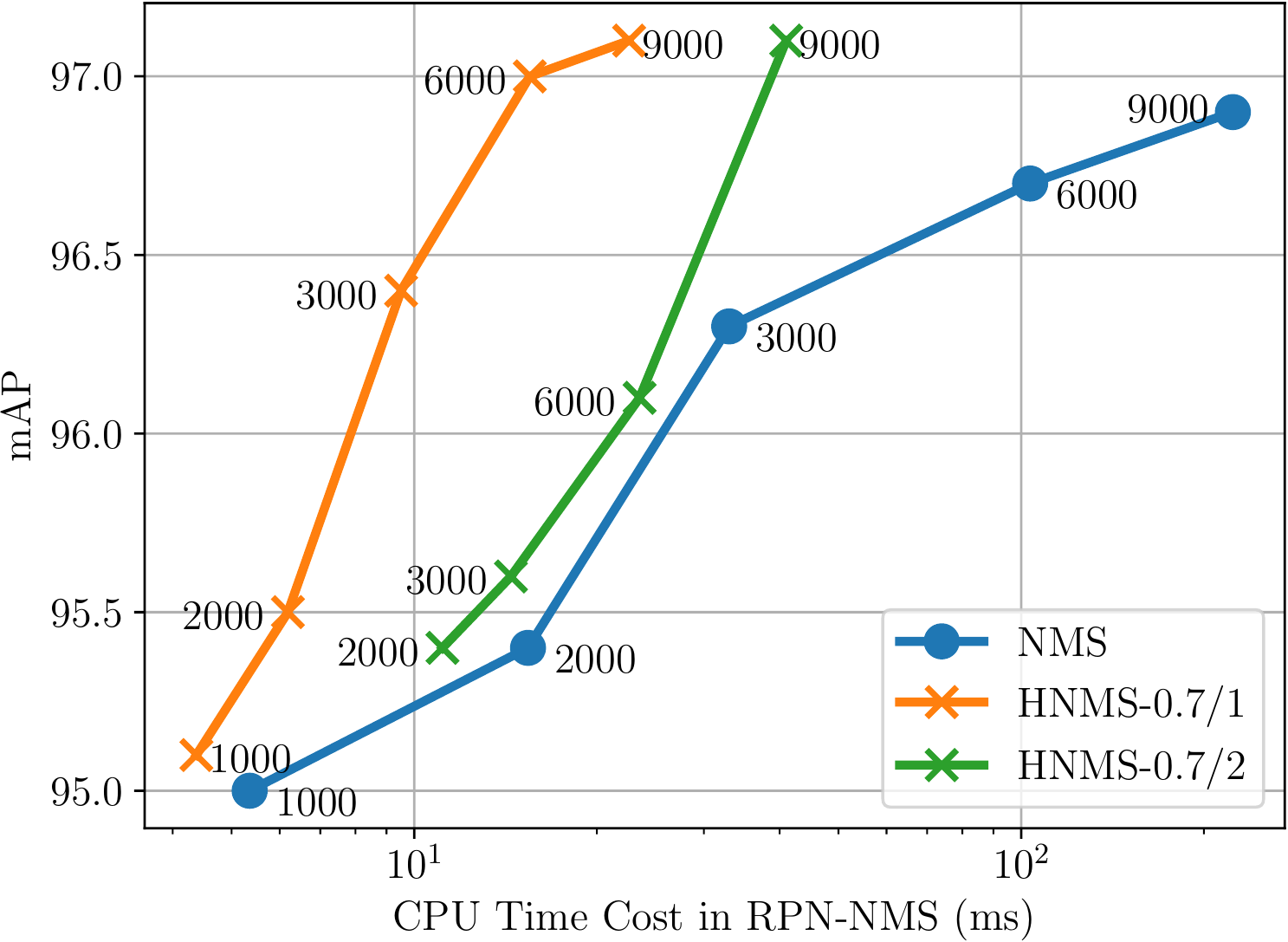}
		& \includegraphics[width=\halffigurelength\linewidth]{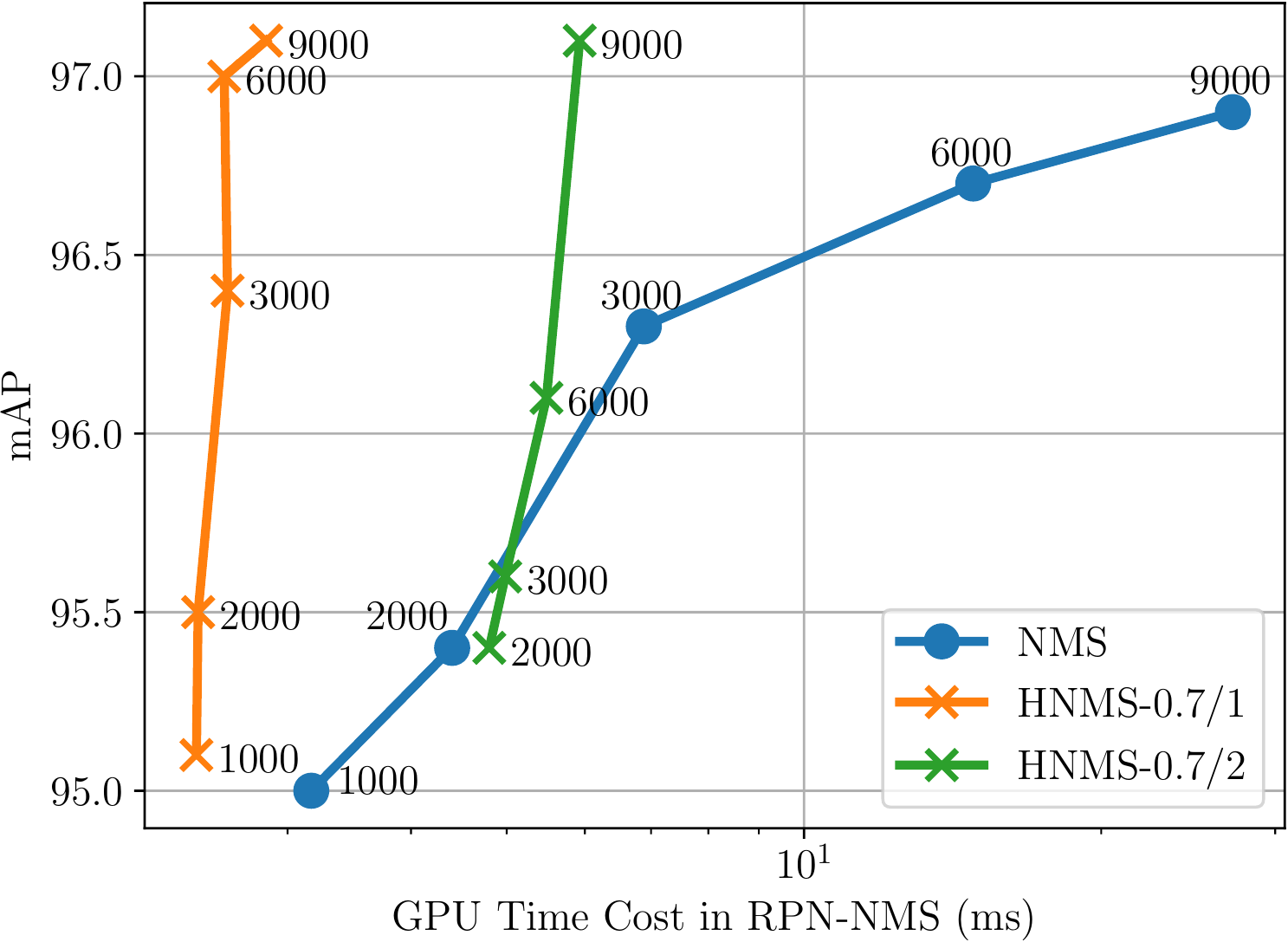} 
	\end{tabular}
	\caption{Accuracy vs time cost with different maximum numbers (shown in the figure)  of boxes for filtering with Faster-RCNN-R50-FPN on CARPK for CPU and GPU. Our approach is denoted as HNMS-$\alpha$/$K$.}
	\label{fig:acc_time}
\end{figure}

\subsection{Results on Two-Stage Detector}
We use Faster R-CNN~\cite{RenHG017} as the test bed to evaluate the performance on two-stage detectors. 
ResNet50 is used as the backbone and feature pyramid network~\cite{LinDGHHB17} is adopted to provide multiple feature maps. 
The network is abbreviated as Faster-RCNN-R50-FPN. 

On CARPK and CrowdHuman, the models are trained with $20$ epochs while on SKU-110K, the model is trained with $80$ epochs.
The initial learning rate are all $0.02$, and is decreased by $10$x twice at $2/3$ and $8/9$ of the total iterations. 
The weight decay is $0.0001$ and the momentum is $0.9$. 
The batch size is $16$ trained on $4$ GPUs. 
During inference, we use at most $P_1$ proposals for NMS on each feature map and keep at most $P_2$ boxes after NMS. At most $P_3$ proposals collected from all feature maps are passed to the RoI head network. 
At most $1$K boxes after RoI Head netowrk are used for evaluation. 
On CARPK and CrowdHuman, we have $P_1, P_2, P_3 = 9K, 2K, 4K$ and on SKU-110K, it is $9K, 1K, 1K$ if these numbers are not explicitly specified. 
The ablation study of the parameters $P_1$ is also presented as follows, which plays an important role in the NMS speed and the accuracy. 

Both RPN and RoI head network apply NMS to suppress the co-located boxes. 
For RPN, the NMS threshold is $0.7$ and we also set 
$\alpha=0.7$ for HNMS. 
The NMS in RoI head network is not altered  since it is fast enough.
For example in CARPK, it takes $3.3$ms and $0.08\%$ of the total time for inference. 

The results are shown in Table \ref{tbl:rpn}. 
As we can see, HNMS can achieve comparable accuracy but with significant less time cost on RPN-NMS  for all three datasets and on both CPU and GPU. 
For instance, on CARPK, we can use one IoUHash function to achieve slightly higher mAP ($97.1\%$ vs $96.9\%$) 
with $7.4$x speed.

When we use more HNMS processes, the accuracy is dropped for CARPK, and increased slightly for SKU-110K and CrowdHuman. 
The reason is that with more filtering, HNMS could suppress more positive boxes, which leads to slight recall drop. 
On the other hand, it can suppress more negative boxes, which improves the precision. 
Another observation is that the time cost of the baseline NMS is not consistent among different datasets. 
The reason is that the complexity ranges from $O(N\log(N) + N)$ to $O(N\log(N) + N^2)$ and thus the time cost is data-dependent. 

\noindent\textbf{Varying $\alpha$}.
Fig. \ref{fig:acc_alpha} shows the accuracy as a function of $\alpha$ with different $K$. With a fixed $K$, the accuracy is normally increased first and 
then decreased with increasing $\alpha$.
The reason is that if $\alpha$ is too small, the hash cell will be quite large, which 
will suppress lots of positive boxes. 
if $\alpha$ is too large, the cell size will become tiny, which fails to suppress enough negative boxes.
In the latter case, the accuracy  can be improved by more HNMS (increasing $K$).
For example in Fig. \ref{fig:acc_alpha}(c) at $\alpha=0.8$, mAP is improved when $K$ is increased from $1$ to $4$.


\noindent\textbf{Varying $P_1$}. 
To reduce the time cost, one can use top fewer boxes (lower $P_1$)
based on objectness in RPN for NMS filtering. 
With different numbers of boxes, we have the result illustrated in Fig.~\ref{fig:acc_time} with our approach denoted as HNMS-$\alpha/K$, and have following observations.
\begin{enumerate}
\item The accuracy can be boosted significantly by simply increasing the number of boxes used for NMS. For example from 1000 to 9000, the accuracy can be improved from 95.0 to 96.9, resulting in nearly 2 points gain on CARPK. This also demonstrates the necessity of more boxes for crowded scene. 
\item With the same time cost, our approach can achieve higher accuracy. 
For example in CPU, NMS obtains $95.4$ mAP with more than $10$ms, but HNMS achieves $96.4$ mAP with less than $10$ms, which results in $1\%$ mAP improvement. 
\item With similar accuracy, HNMS can achieve much faster speed, which is consistent with the results in Table~\ref{tbl:rpn}.
\item In GPU, time cost of our approach is less sensitive to the number of boxes. For HNMS-0.7/1, the time cost increases from $2.4$ms slightly to $2.9$ms when the number of boxes is from $1000$ to $9000$.
This is because our approach is more friendly in parallel computing. 
\end{enumerate}

\begin{table}[!t]
\centering
\caption{Experimental results with Retina-R50-FPN. The columns of HNMS, NMS and Total represent the time cost in ms. Speed is for the total cost relative to NMS.}
\label{tbl:retina}
\scalebox{\scaletable}{
\begin{tabular}{c}
	(a) CARPK
	\\
\begin{tabular}{c@{~~}c@{~~}c@{~~~}c@{~}c@{~}c@{~}c@{~~~}c@{~}c@{~}c@{~}c}
\toprule
 &  &  & \multicolumn{4}{c}{CPU} & \multicolumn{4}{c}{GPU}\\
\cmidrule(lr){4-7}
\cmidrule(lr){8-11}
 & $K$ & mAP & HNMS & NMS & Speed & Total & HNMS & NMS & Speed & Total\\
\midrule
NMS &  & 94.8 &  & 260.9 & 1x & 2071.1 &  & 134.7 & 1x & 250.9\\
\midrule
 & 1 & \textbf{95.3} & 23.6 & 194.2 & \textbf{1.2x} & 1761.9 & 0.9 & 33.1 & \textbf{4.0x} & 149.1\\
HNMS & 2 & \textbf{95.3} & 34.0 & 161.7 & \textbf{1.3x} & 1762.6 & 1.4 & 18.4 & \textbf{6.8x} & 121.9\\
+NMS & 3 & 95.2 & 46.1 & 140.4 & 1.4x & 1794.6 & 2.5 & 19.5 & 6.1x & 142.1\\
 & 4 & 95.2 & 51.4 & 122.4 & 1.5x & 1732.6 & 3.0 & 14.6 & 7.7x & 131.3\\
\bottomrule
\end{tabular}
\\
\\
(b) SKU-110K
\\
\begin{tabular}{c@{~~}c@{~~}c@{~~~}c@{~}c@{~}c@{~}c@{~~~}c@{~}c@{~}c@{~}c}
\toprule
 &  &  & \multicolumn{4}{c}{CPU} & \multicolumn{4}{c}{GPU}\\
\cmidrule(lr){4-7}
\cmidrule(lr){8-11}
 & $K$ & mAP & HNMS & NMS & Speed & Total & HNMS & NMS & Speed & Total\\
\midrule
NMS &  & \textbf{91.6} &  & 185.2 & \textbf{1x} & 1640.9 &  & 121.1 & \textbf{1x} & 217.9\\
\midrule
 & 1 & \textbf{91.6} & 19.4 & 124.5 & \textbf{1.3x} & 1576.3 & 0.7 & 18.8 & \textbf{6.2x} & 115.8\\
HNMS & 2 & \textbf{91.6} & 33.9 & 104.2 & \textbf{1.3x} & 1567.6 & 1.4 & 14.0 & \textbf{7.9x} & 111.3\\
+NMS & 3 & 91.5 & 42.7 & 88.2 & 1.4x & 1571.8 & 2.1 & 11.3 & 9.1x & 109.0\\
 & 4 & 91.5 & 50.3 & 81.0 & 1.4x & 1531.5 & 2.8 & 10.9 & 8.9x & 111.4\\
\bottomrule
\end{tabular}
\end{tabular}
}
\end{table}

\begin{figure}[t]
	\centering
	\begin{tabular}{c@{}c@{}c@{}c}
		\includegraphics[width=\fourthfigurelength\linewidth]{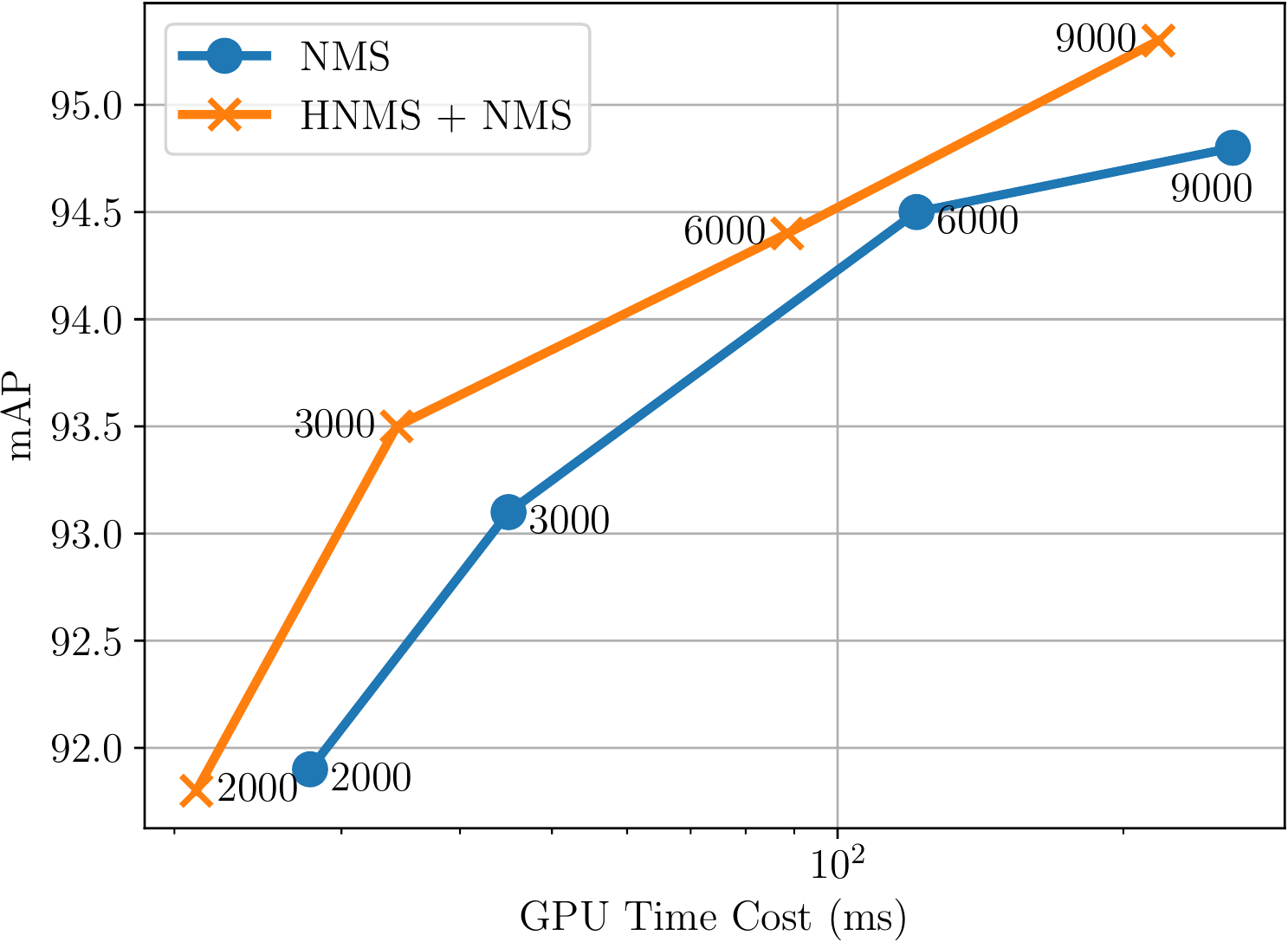} & 
		\includegraphics[width=\fourthfigurelength\linewidth]{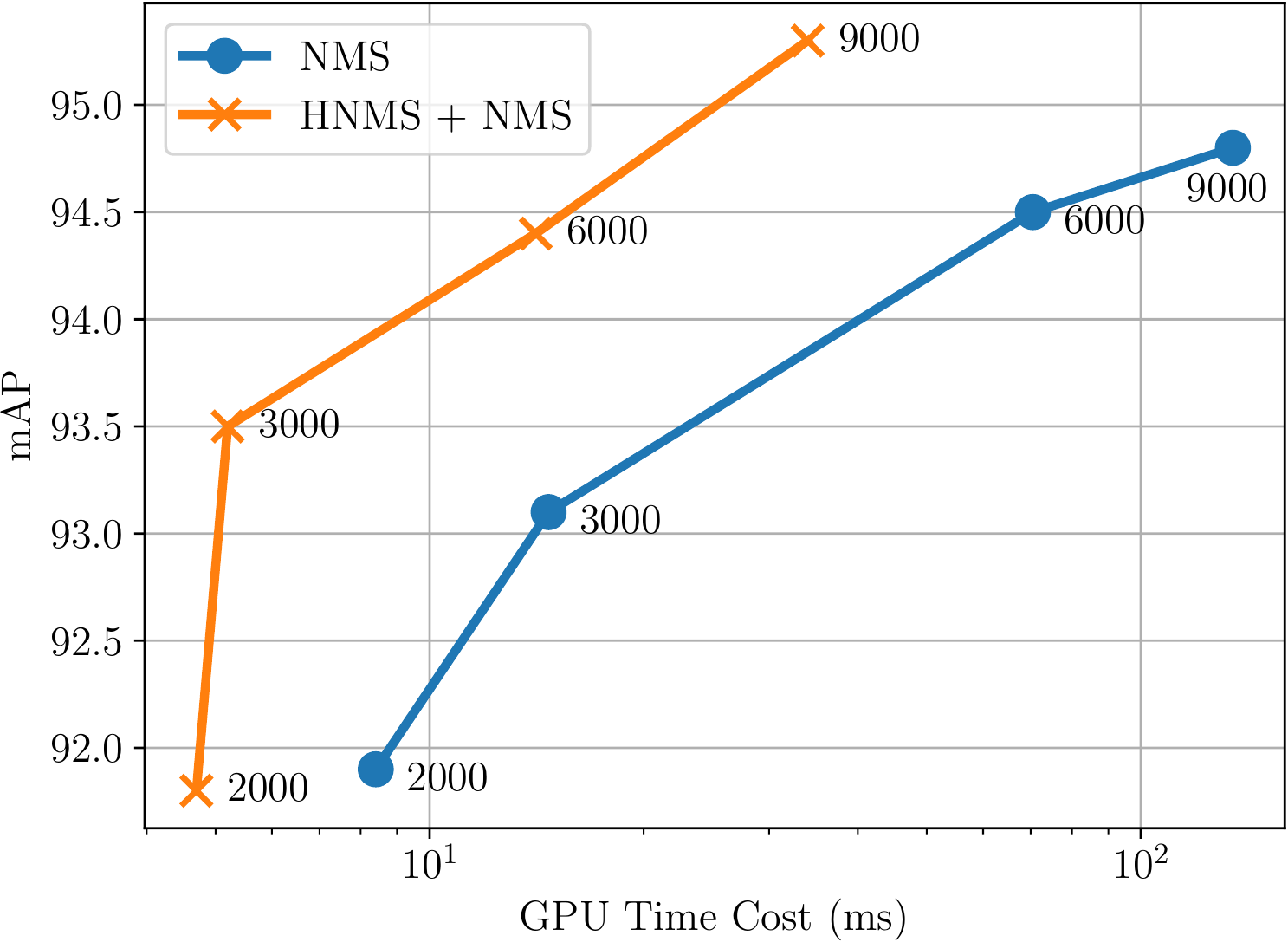} &
		\includegraphics[width=\fourthfigurelength\linewidth]{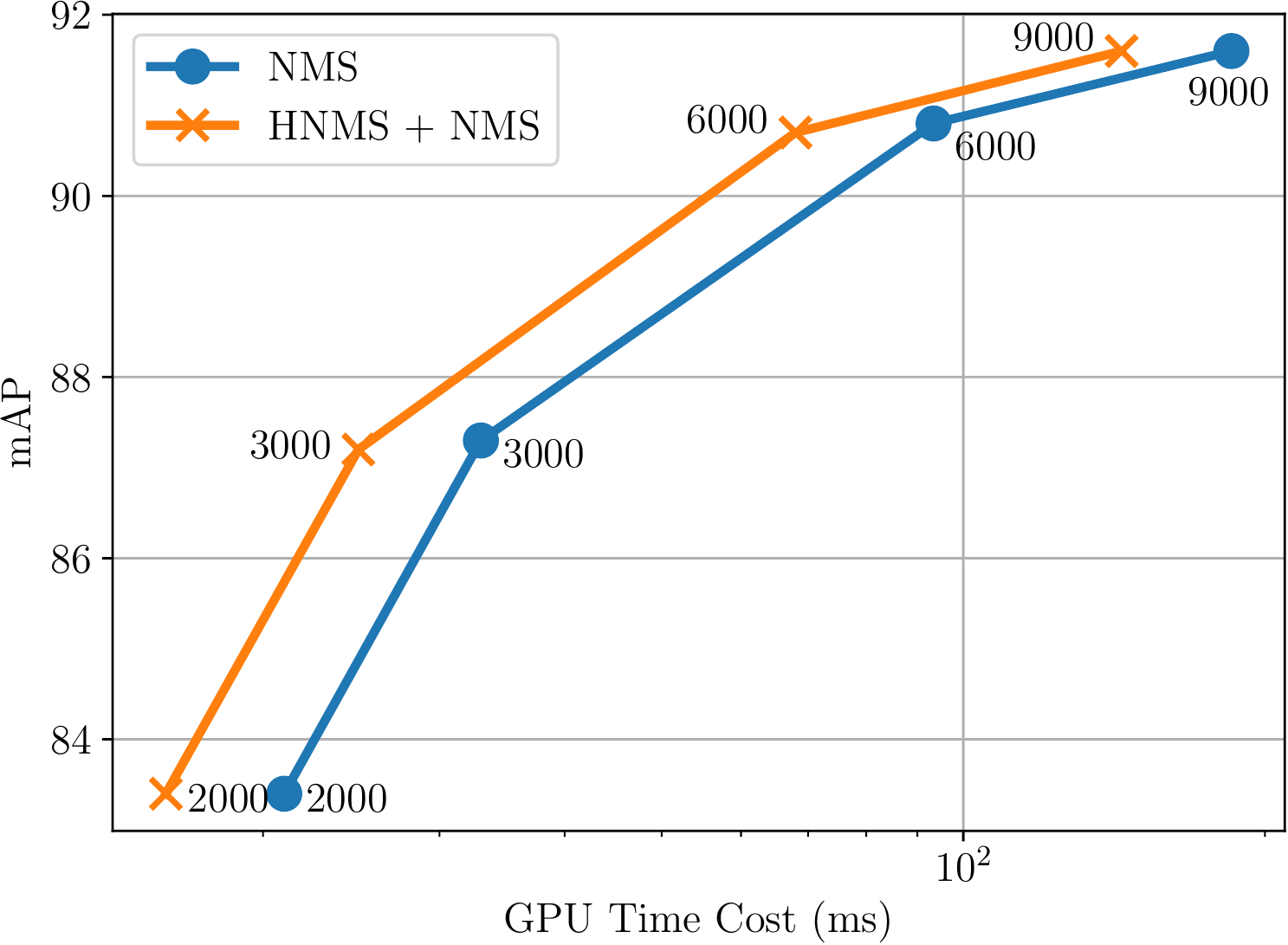} &
		\includegraphics[width=\fourthfigurelength\linewidth]{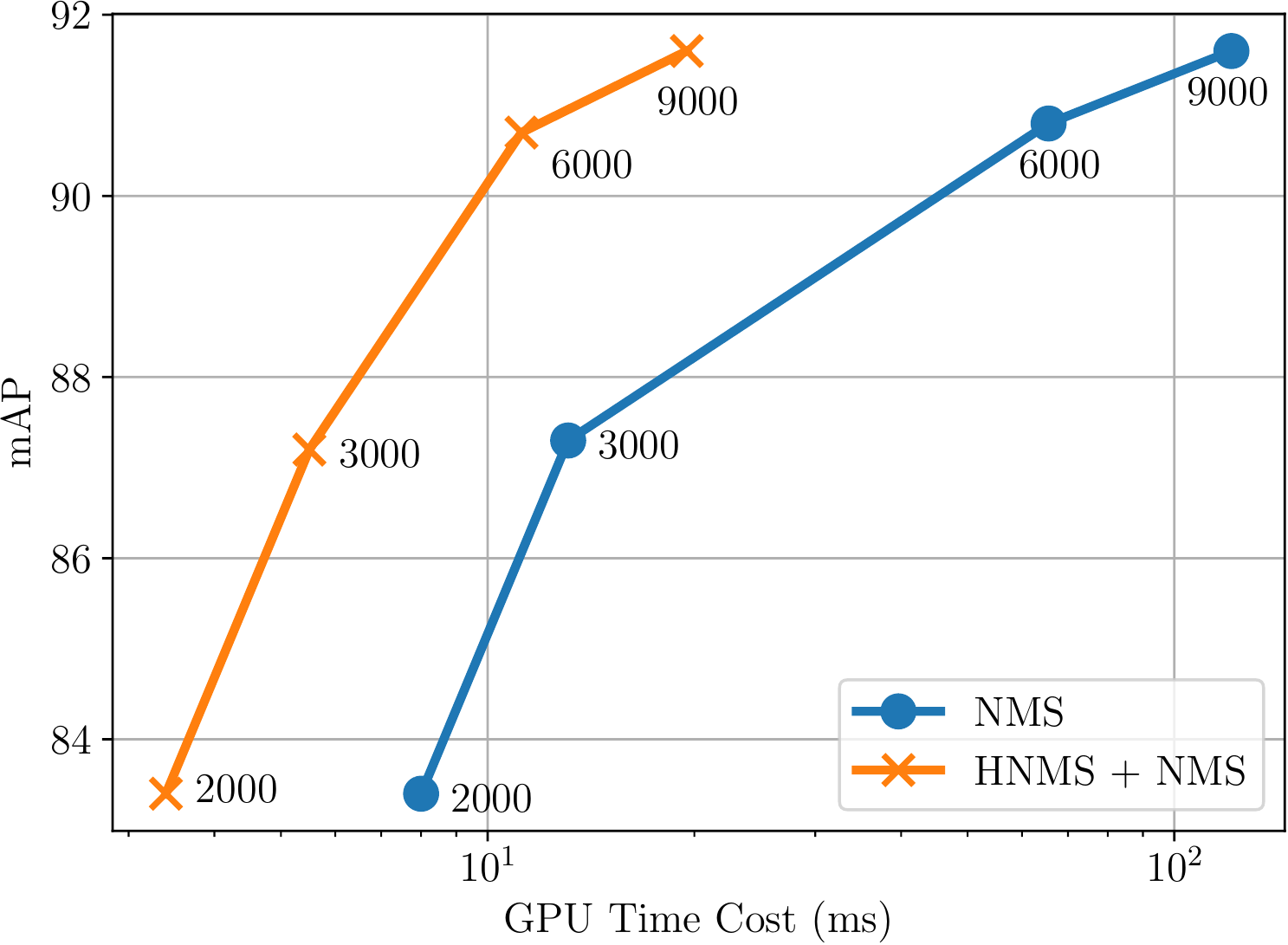} 
		\\
		\multicolumn{2}{c}{\scriptsize{CARPK}} & \multicolumn{2}{c}{\scriptsize{SKU-110K}}
	\end{tabular}
	\caption{Accuracy vs GPU time cost with different maximum numbers of boxes for suppression filtering in Retina-R50-FPN. HNMS is as a pre-filter before applying the NMS.}
	\label{fig:retina_acc_time}
\end{figure}


\subsection{Results on One-Stage Detector}
RetinaNet~\cite{LinDGHHB17} is used as a representative one-stage detector
and Resnet50 with FPN is as the backbone. The model is short as
Retina-R50-FPN.
On CARPK and SKU-110K, the model is trained with $40$ epochs, while on CrowdHuman, it is $20$ epochs. The initial learning rate is $0.01$ for CARPK, $0.005$ for SKU-110K and $0.02$ for CrowdHuman. The learning rate is decreased by $10$x twice at $2/3$ and $8/9$ of the total iterations.
The weight decay is $0.0001$, and the momentum is $0.9$. The batch size is $16$ trained on $4$ GPUs. 
For NMS, we use at most $P_1 = 9K$ boxes by default and use at most $1$K final boxes for evaluation. 
The NMS threshold is $0.5$. 

As discussed in Sec.~\ref{sec:discussion}, we use HNMS as a pre-filter before applying NMS. To reduce suppressing positive boxes, we use a higher $\alpha=0.73$. The corresponding lower bound is 
$0.502$, slightly higher than the NMS threshold ($0.5$). 
The results are shown in Table~\ref{tbl:retina} with different $K$
for CARPK and SKU-110K. The results on CrowdHuman can be found in Appendix \ref{sec:supp_table_mising_crowd}.
When $K = 1$, the accuracy of HNMS is consistently comparable (or slightly better) with the baseline, but with higher speed on the suppression component.
For example on CARPK with GPU, the speed-up is $4$x ($6.8$x for $K = 2$) with slightly better accuracy. 
On SKU-110K, the speed-up is $6.2$x ($7.9$x for $K = 2$) with the same accuracy. 

When we increase $K$, more time is spent for pre-filtering, 
but fewer boxes are passed to NMS. 
The accuracy is penalized because more positive boxes could be removed as discussed in Sec.~\ref{sec:discussion}.
For example on SKU-110K with GPU, when $K$ changes from $1$ to $4$, the time cost of pre-filtering takes $0.7$ms to $2.8$ms. The time cost of NMS is decreased
from $18.8$ms to $14.0$ms, and the overall speed is improved from $6.2$x to around $9$x. The accuracy stays the same at $K = 1, 2$, and is slightly dropped by $0.1$ point at $K = 3, 4$.

\noindent\textbf{Varying $P_1$.}
Fig.~\ref{fig:retina_acc_time} shows the accuracy as a function of time cost 
in NMS or HNMS+NMS when we vary the maximum numbers of boxes used for filtering with $K = 1$. With comparable accuracy, the time cost with HNMS as pre-filter always reduces dramatically, especially for GPU.

One alternative is to replace NMS with SoftNMS~\cite{BodlaSCD17}, and use our HNMS for pre-filtering. We leave this result in Appendix \ref{sec:softnms}.

\section{Conclusion}
We studied the efficiency problem of NMS in object detection and proposed a Hashing-based NMS algorithm to improve the speed. The key idea is to hash
each bounding box to a discrete cell and suppress the boxes with smaller 
confidences within each cell. To implement this, we proposed a novel IoUHash function, 
which guarantees the closeness of the boxes by a lower IoU bound. 
Comprehensive experiments were conducted to verify the significant speed improvement 
with comparable accuracy.

\clearpage
%
%
\bibliographystyle{splncs04}
\bibliography{bib}

\clearpage

\appendix
{\LARGE{\textbf{Appendix}}}

\section{IoU Lower Bound}\label{sec:iou_lower_bound}
In Sec. 3.1 (main paper), we discussed how the IoU lower bound (given two boxes are hashed to the same cell) is calculated. Alg. \ref{alg:low} summarizes the details. 

To derive this algorithm, we implicitly relied on the conclusions 
in Theorem~\ref{thm:relation} and Theorem~\ref{thm:boundary}. Here, we present the proof. 

\begin{theorem}
	\label{thm:relation}
	With Eqn. 15 and Eqn. 14 (main paper), IoU is unrelated with $b_x$, $b_y$, $W_0$, $H_0$, $i$, $j$, $m$, $n$, 
but only depends on $\alpha$,  $i_k$, $j_k$, $m_k$ and $n_k$.
\end{theorem}
\begin{proof}
As demonstrated in the paper, the intersection is larger than 0 since Eqn. 14 (main paper) holds. 
Thus, we can write the intersection as
\begin{align}
	I = F(r - l) F(b - t) = (r -l)(b - t),
\end{align}
where $r$, $l$, $b$, $t$ are defined in Eqn. 10 (main paper) and Eqn. 11 (main paper), $F(x)=x$ if $x \ge 0$ and 0, otherwise. 
Substituting Eqn. 10 (main) and Eqn. 15 (main paper), we have
\begin{align}
r - l & = \min(x_1 + \frac{1}{2}w_1, x_2 + \frac{1}{2}w_2) - \max(x_1 - 
\frac{1}{2}w_1, x_2 - \frac{1}{2}w_2)
\label{eqn:r_minus_l}
\\
(x_k \pm \frac{1}{2}w_k)
& = b_x \delta_i + (m + m_1) \delta_i \pm \frac{W_0}{2\alpha^{i + i_k}} \\
& = b_x \delta_i + m\delta_i + m_1 \delta_i \pm \frac{W_0}{2\alpha^{i + i_k}},
k \in \{1, 2\}
\end{align}
The item of $(b_x \delta_i + m\delta_i)$ is unrelated with $k$, and 
thus it can be removed for Eqn.~\ref{eqn:r_minus_l}. 
With the definition of $\delta_i$ in Eqn. 4 (main paper), we have
\begin{align}
r - l & = \frac{W_0}{\alpha^{i}}F_w(i_1, i_2, m_1, m_2) \\
F_w() & \triangleq
\min(m_1 + \frac{1}{2\alpha^{i_1}}, m_2 + \frac{1}{2\alpha^{i_2}}) -
\max(m_1 - \frac{1}{2\alpha^{i_1}}, m_2 - \frac{1}{2\alpha^{i_2}})
\label{eqn:fw}
\end{align}
where $F_w()$ is a function after we extract $W_0 / \alpha^i$ and  does not depend on $W_0$ and $i$.
Similarly, we have
\begin{align}
b - t = \frac{H_0}{\alpha^{j}}F_h(j_1, j_2, n_1, n_2).
\end{align}
The area of the two boxes are 
\begin{align}
A_k = \frac{W_0H_0}{\alpha^{i + j + i_k + j_k}}, k \in \{1, 2\}.
\end{align}
Thus, IoU can be calculated by
\begin{align}
\textrm{IoU} & = \frac{I}{A_1 + A_2 - I} = \frac{1}{\frac{A_1 + A_2}{I} - 1} \\
& = \frac{}{
\frac{1/\alpha^{i_1+j_1} + 1 / \alpha^{i_2 + j_2}}{F_w(i_1, i_2, m_1, m_2)F_h(j_1, j_2, n_1, n_2)} - 1
},
\label{eqn:iou}
\end{align}
which demonstrates that IoU has no relation with the cell index, but only depends on the 
offsets to the cell center. 
\end{proof}

\begin{lemma}
	The minimum IoU is located at the boundary of $m_1$ (equal to -0.5 or 0.5 since the range is from -0.5 to 0.5 as in Eqn. 15 of the main paper) given all other variables fixed. 
	\label{lm:m}
\end{lemma}
\begin{proof}
Since $A_{k}$, $t - b$ has no relation with $m_1$, the lemma is equivalent to prove that the minimum of $F_w$ in Eqn.~\ref{eqn:fw} is located at
the boundary of $m_1$.
The first min operation in $F_w$ is a concave  function of $m_1$, and 
the second of negative max operation is also a concave function of $m_1$. 
Thus, $F_w$ is concave with $m_1$, which concludes that the minimum value
must be at the boundary of $m_1$. 
\end{proof}

\begin{lemma}
	The minimum IoU is located at the boundary of $i_1$ (equal to -0.5 or 0.5) given all other variables fixed. 
	\label{lm:i}
\end{lemma}
\begin{proof}
Rather than starting from Eqn.~\ref{eqn:iou}, we study each position relationship between two overlapped boxes. 
Fig~\ref{fig:position} enumerates all relations. The idea is to verify for each position relationship, IoU is the smallest if $w_1$ (parameterized by $i_1$) is at the boundary with all other variables fixed. 
\begin{figure}[t!]
	\centering
\includegraphics[width=0.8\linewidth]{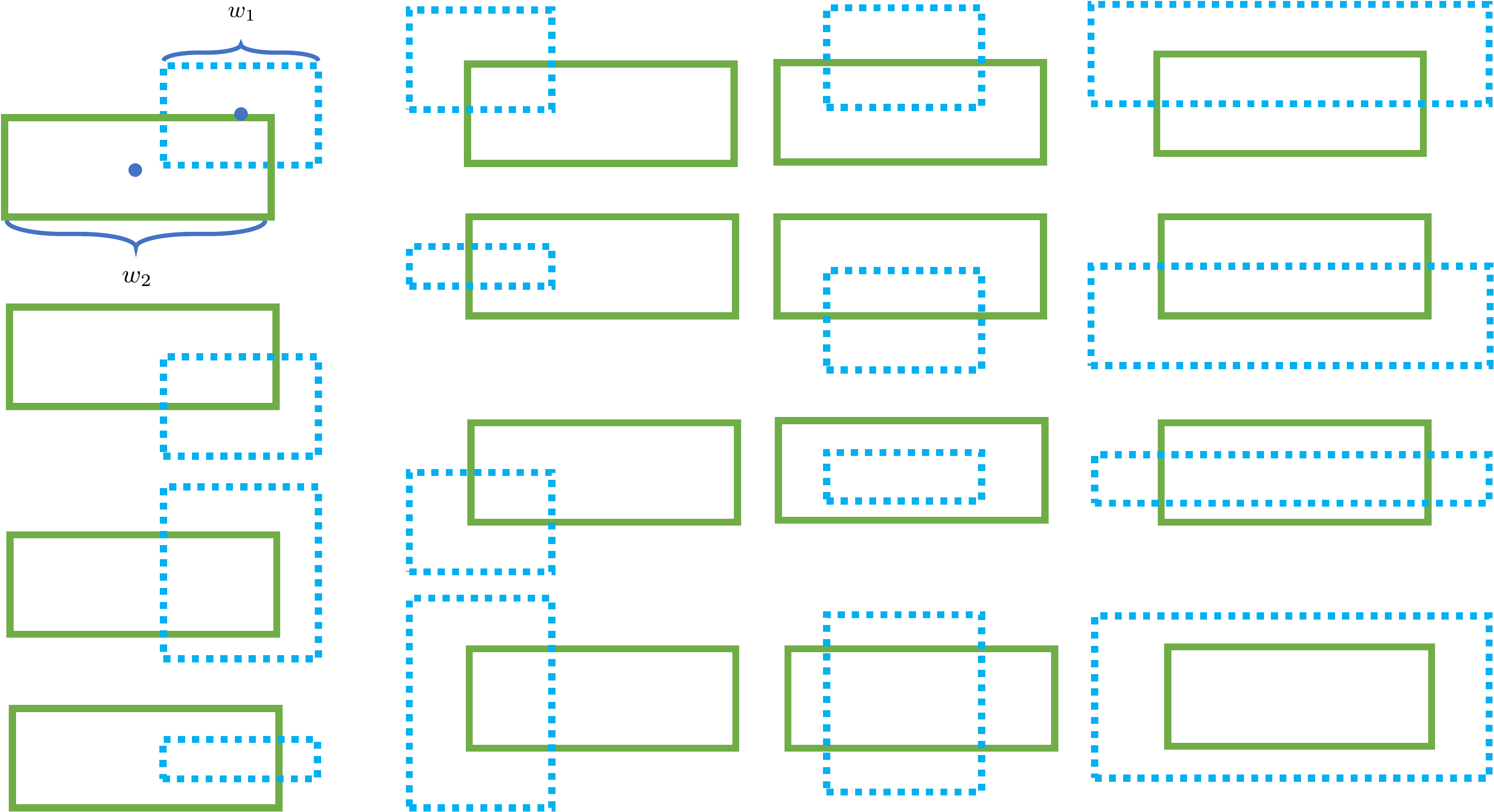}
\caption{Different positions between two overlapped boxes. Assume that the position of the green box is fixed, and the blue box can be floating anywhere as long as they are overlapped. }
\label{fig:position}
\end{figure}

Taking the top-left one as an example, we can write the intersection as 
\begin{align}
I  = (\frac{1}{2}w_1 + \frac{1}{2}w_2 - |x_1 - x_2|)
(\frac{1}{2}h_1 + \frac{1}{2}h_2 - |y_1 - y_2|)
\end{align}
which is a linear function of $w_1$. Since the area is also linear with $w_1$, IoU is monotonous with $w_1$.
If it is monotonously increasing, IoU achieves the smallest if $w_1$ 
is the smallest or $i_1$ is equal to -0.5. If it is monotonously decreasing, 
we can increase $i_1$ or $w_1$ until the relationship becomes the top-right of Fig.~\ref{fig:position}, where IoU is even smaller if the width is even wider. Thus, IoU achieves the smallest when $w_1$ is the largest or $i_1$ equals $0.5$.
Verifying all other relations, we can conclude that $i_1$ should be in the boundary if IoU is the smallest. 
\end{proof}

\begin{theorem}
	\label{thm:boundary}
	Under Eqn. 15 and Eqn. 14 in the main paper, the minimum IoU must reside in one of the boundaries, where $i_k$, $j_k$, $m_k$, and $n_k$ equals -$0.5$ or $0.5$. 
\end{theorem}

\begin{proof}
Assume that $i_k^*$, $j_k^*$, $m_k^*$ and $n_k^*$ gives the minimum IoU, and 
at least one of the variables is not at the boundary.
Given the Lemma~\ref{lm:m} and Lemma~\ref{lm:i}, we can conclude IoU can be lower if that variable goes to the boundary with other variables fixed. 
Thus, all those variables must be in the boundary if IoU is smallest.
\end{proof}

\begin{algorithm}[t]
	\caption{Lower IoU Bound for IoUHash}
	\label{alg:low}
	\begin{algorithmic}[1]
		\REQUIRE
		$\alpha$
		\ENSURE
		lower IoU bound
		\IF{Eqn. 14 does not hold}
		\STATE{return 0}
		\ENDIF
		\STATE{$\text{min}\_{\text{iou}}$ = 1}
		\STATE{$W_0, H_0, b_x, b_y$ = 1, 1, 0, 0}
		\FOR{each $\{i_k, j_k, m_k, n_k|k = 1, 2\}$ in $\{-0.5, 0.5\}^8$}
		\STATE {Derive $\{w_k, h_k, x_k, y_k| k = 1, 2\}$ from Eqn. 15}
		\STATE{Calculate iou}
		\IF{$\text{min}\_\text{iou} > \text{iou}$}
		\STATE{min\_iou = iou}
		\ENDIF
		\ENDFOR
		\STATE{return min\_iou}
	\end{algorithmic}
\end{algorithm}

\section{GPU Implementation}\label{sec:gpu_impl}
Sec. 3 (main paper) briefly discussed the GPU implementation. 
Here we present more details, as shown in Fig. \ref{fig:iou}.

\begin{figure}[t!]
	\centering
\includegraphics[width=0.99\linewidth]{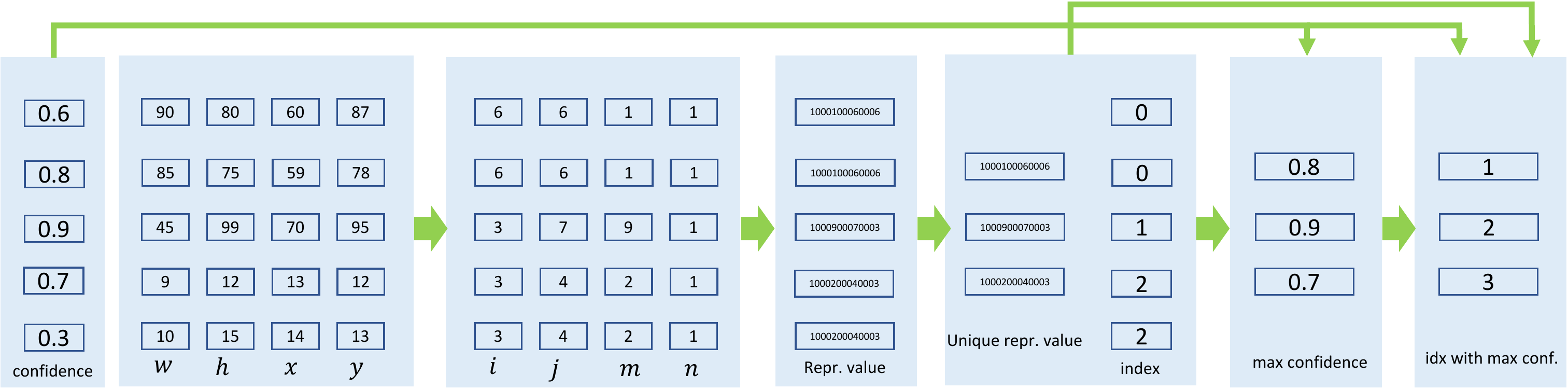}
\caption{Illustration of GPU Implementation. Each bounding box is first hashed to the code, which is further mapped to a
representative value. Then, a unique representative code
and the inverse index for each box is calculated. Third, the maximum confidence with each cell is calculated based on atomic operation of max. Last, the final index is derived based on atomic operation of compare-and-swap. }
\label{fig:iou}
\end{figure}

The first step is to hash each box into the hash index ($i, j, m, n$), and then map it to a \textit{representative value} by 
\begin{align}
c(i, j, m, n) = i + j * 10^4 + m * 10^8 + n * 10^{12}.
\label{eqn:c}
\end{align}
The function can be any one as long as the code $(i, j, m, n)$ is deterministic 
by the representative value, which helps to identify which boxes shares the same code. Although Eqn~\ref{eqn:c}
does not satisfy the requirement theoretically, 
the function
works in practice since the image size is normally limited (less than $2000\times 2000$).

Next, we leverage the implementation of
torch.unique\footnote{https://pytorch.org/docs/master/torch.html\#torch.unique}
to get the unique representative codes and the reverse index of each box to the unique code. 

Third, we calculate the maximum confidence value within the same hash cell.
Specifically, each bounding box is scanned one by one (not the unique representative code), and an array is allocated to record the largest confidence (initialized as 0).
If the box's confidence is higher than the recorded largest confidence, then replace it. 
Since multiple GPU cores could modify the value at the same address, we use the atomic operation of atomicMax\footnote{https://docs.nvidia.com/cuda/cuda-c-programming-guide/index.html\#atomicmax}
in CUDA library to compare and replace. 
The funciton of atomicMax only accepts the integer input, so each 
floating-valued confidence score $f$ is converted to an integer by $f\times 10^6 + 10^5$. We will discuss the usage of addition with $10^5$ in the last step.
Since the confidence is within $0$ to $1$, this conversion 
is enough in practice. 

Last, we find the index with the maximum confidence score for 
each unique code. This process is implemented by 
scanning each bounding box  
based on the atomic operation of atomicCAS\footnote{https://docs.nvidia.com/cuda/cuda-c-programming-guide/index.html\#atomiccas}.
That is, for each bounding box, swap its index if the box's integer-converted confidence equals the maximum one in the array. 
Note, after the index is swapped, the array stores the index rather than the integer-converted confidence. To make sure the index is not swapped again by other boxes, we make the integer-converted confidence at least $10^5$.

\section{Experiments}
\subsection{VOC and COCO}\label{sec:voc_coco}
Pascal VOC~\cite{everingham2015pascal} and COCO~\cite{lin2014microsoft} are two widely-used datasets for object
detection.
On average, each image has 3 boxes in VOC and 7 boxes for COCO.
Since the object density is small, it is not beneficial to feed more boxes into 
NMS. On VOC, we trained a model of Faster-RCNN-R34-FPN 
with 9000 iterations and achieved consistent 77.2\% mAP for $P_1$\footnote{Recall that $P_1$ is the maximum number of boxes in each feature map (5 feature maps in total) used for NMS in RPN.}
ranging from 1000 to 9000. 
On COCO, we trained a model of Faster-RCNN-R50-FPN with 180000
iterations and achieved consistent 37.4\% mAP$_{0.5:0.95}$
for $P_1$ ranging from 1000 to 9000. 
The accuracy on COCO is based on the average mAP over IoU threshold 0.5:0.05:0.95 as commonly adopted in this dataset, and all other accuracies are based on mAP at 0.5 as described in the main paper. 

Regarding the time cost at $P_1 = 1000$ for CPU, 
NMS takes $5.5 / 1473.6 = 0.4\%$
of the total cost on VOC and $6.5 / 1998.5 = 0.3\%$ on COCO. 
For GPU, it is $2.3 / 47.9 = 4.8\%$ on VOC and $2.7 / 67.1 = 4.0\%$ on COCO.
Due to the low time cost, we mainly focus on the crowded scenarios as in the main paper, e.g. in CARPK~\cite{HsiehLH17} with an average of 103 objects for each image. 

\subsection{Supplementary for Table 2}\label{sec:supp_table_mising_crowd}
In Table 2 of the main paper, we showed results on CARPK and SKU-100K. 
Fig.~\ref{tbl:crowdhuman} shows the results on CrowdHuman
and we have similar observations: comparable accuracy but higher speed, especially for GPU.

\begin{table}[t!]
	\centering
\caption{Experimental results on CrowdHuman with Retina-R50-FPN. The columns of HNMS, NMS and Total represent the time cost in ms. Speed is for the total cost relative to NMS.}
\label{tbl:crowdhuman}
\begin{tabular}{c@{~~}c@{~~}c@{~~~}c@{~}c@{~}c@{~}c@{~~~}c@{~}c@{~}c@{~}c}
\toprule
 &  &  & \multicolumn{4}{c}{CPU} & \multicolumn{4}{c}{GPU}\\
\cmidrule(lr){4-7}
\cmidrule(lr){8-11}
 & $K$ & mAP & HNMS & NMS & Speed & Total & HNMS & NMS & Speed & Total\\
\midrule
NMS &  & 77.7 &  & 327.8 & 1x & 1870.9 &  & 143.9 & 1x & 260.5\\
\midrule
 & 1 & \textbf{77.8} & 25.0 & 268.7 & \textbf{1.1x} & 1653.4 & 0.9 & 65.1 & \textbf{2.2x} & 182.1\\
HNMS & 2 & 77.6 & 42.2 & 234.5 & 1.2x & 1837.4 & 1.6 & 39.4 & 3.5x & 151.3\\
+NMS & 3 & 77.5 & 56.8 & 216.4 & 1.2x & 1858.9 & 2.3 & 29.4 & 4.5x & 139.1\\
 & 4 & 77.4 & 66.2 & 193.5 & 1.3x & 1841.9 & 3.2 & 24.3 & 5.2x & 142.2\\
\bottomrule
\end{tabular}
\end{table}

\begin{table}[t!]
\centering
\caption{Experiment results with Retina-R50-FPN for SoftNMS. The columns of HNMS, NMS and Total represent the CPU time cost in ms. Speed is for the total cost by HNMS and NMS.}
\label{tbl:soft_retina}
\begin{tabular}{c}
	(a) CARPK \\
\begin{tabular}{c@{~}c@{~}c@{~~}c@{~~}c@{~~}c@{~~}c}
\toprule
 & $K$ & mAP & HNMS & SoftNMS & Speed & Total\\
\midrule
NMS &  & 95.3 &  & $40710.5\pm 4402.8$ & 1x & $42980.8\pm 6686.9$\\
\midrule
\multirow{4}{*}{HNMS} & 1 & 95.3 & $22.6\pm 5.4$ & $10160.4\pm 2582.2$ & 4.00x & $11878.1\pm 2591.7$\\
 & 2 & \textbf{95.4} & $31.8\pm 6.3$ & $7210.5\pm 1925.8$ & \textbf{5.62x} & $8901.7\pm 1939.2$\\
 & 3 & \textbf{95.4} & $39.5\pm 8.1$ & $5451.7\pm 1565.6$ & \textbf{7.41x} & $7133.3\pm 1573.8$\\
 & 4 & \textbf{95.4} & $49.4\pm 11.2$ & $4701.4\pm 1372.1$ & \textbf{8.57x} & $6415.1\pm 1402.6$\\
\bottomrule
\end{tabular} \\
\\
(b) SKU-110K \\
\begin{tabular}{c@{~}c@{~}c@{~~}c@{~~}c@{~~}c@{~~}c}
\toprule
 & $K$ & mAP & HNMS & SoftNMS & Speed & Total\\
\midrule
NMS &  & 91.9 &  & $37673.2\pm 2564.3$ & 1x & $39260.5\pm 2692.3$\\
\midrule
\multirow{4}{*}{HNMS} & 1 & 91.9 & $21.8\pm 3.5$ & $11042.8\pm 1837.5$ & 3.40x & $12639.3\pm 1877.0$\\
 & 2 & \textbf{92.0} & $30.3\pm 5.5$ & $6902.1\pm 1144.8$ & \textbf{5.43x} & $8434.5\pm 1214.3$\\
 & 3 & \textbf{92.0} & $38.1\pm 6.5$ & $5524.2\pm 943.4$ & \textbf{6.77x} & $7038.9\pm 1011.7$\\
 & 4 & \textbf{92.0} & $45.5\pm 7.0$ & $4697.9\pm 899.0$ & \textbf{7.94x} & $6258.0\pm 988.1$\\
\bottomrule
\end{tabular}\\
\\
(c) CrowdHuman \\
\begin{tabular}{c@{~}c@{~}c@{~~}c@{~~}c@{~~}c@{~~}c}
\toprule
 & $K$ & mAP & HNMS & SoftNMS & Speed & Total\\
\midrule
NMS &  & \textbf{78.6} &  & $44272.1\pm 3350.4$ & \textbf{1x} & $45821.6\pm 3418.3$\\
\midrule
\multirow{4}{*}{HNMS} & 1 & \textbf{78.6} & $26.7\pm 6.8$ & $17549.4\pm 2625.1$ & \textbf{2.52x} & $19295.2\pm 2706.6$\\
 & 2 & \textbf{78.6} & $39.0\pm 7.4$ & $11474.2\pm 2039.7$ & \textbf{3.85x} & $13125.2\pm 2053.0$\\
 & 3 & \textbf{78.6} & $51.5\pm 8.8$ & $9234.0\pm 1838.6$ & \textbf{4.77x} & $10937.6\pm 1863.9$\\
 & 4 & \textbf{78.6} & $65.7\pm 35.9$ & $7940.9\pm 1713.3$ & \textbf{5.53x} & $9717.2\pm 1791.1$\\
\bottomrule
\end{tabular}
\end{tabular}
\end{table}

\begin{figure}[t!]
	\centering
	\begin{tabular}{ccc}
	\includegraphics[width=0.32\linewidth]{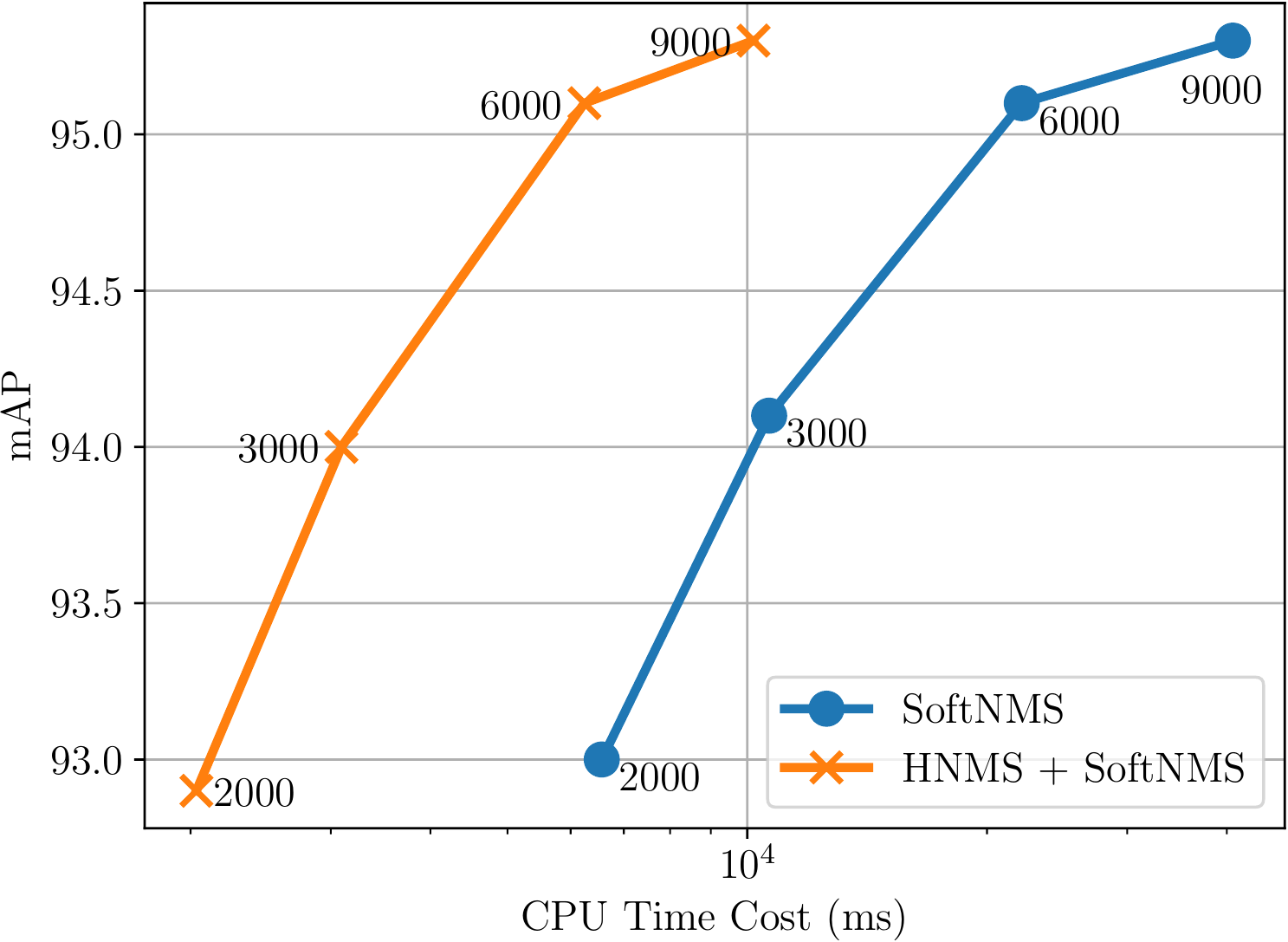} &
	\includegraphics[width=0.32\linewidth]{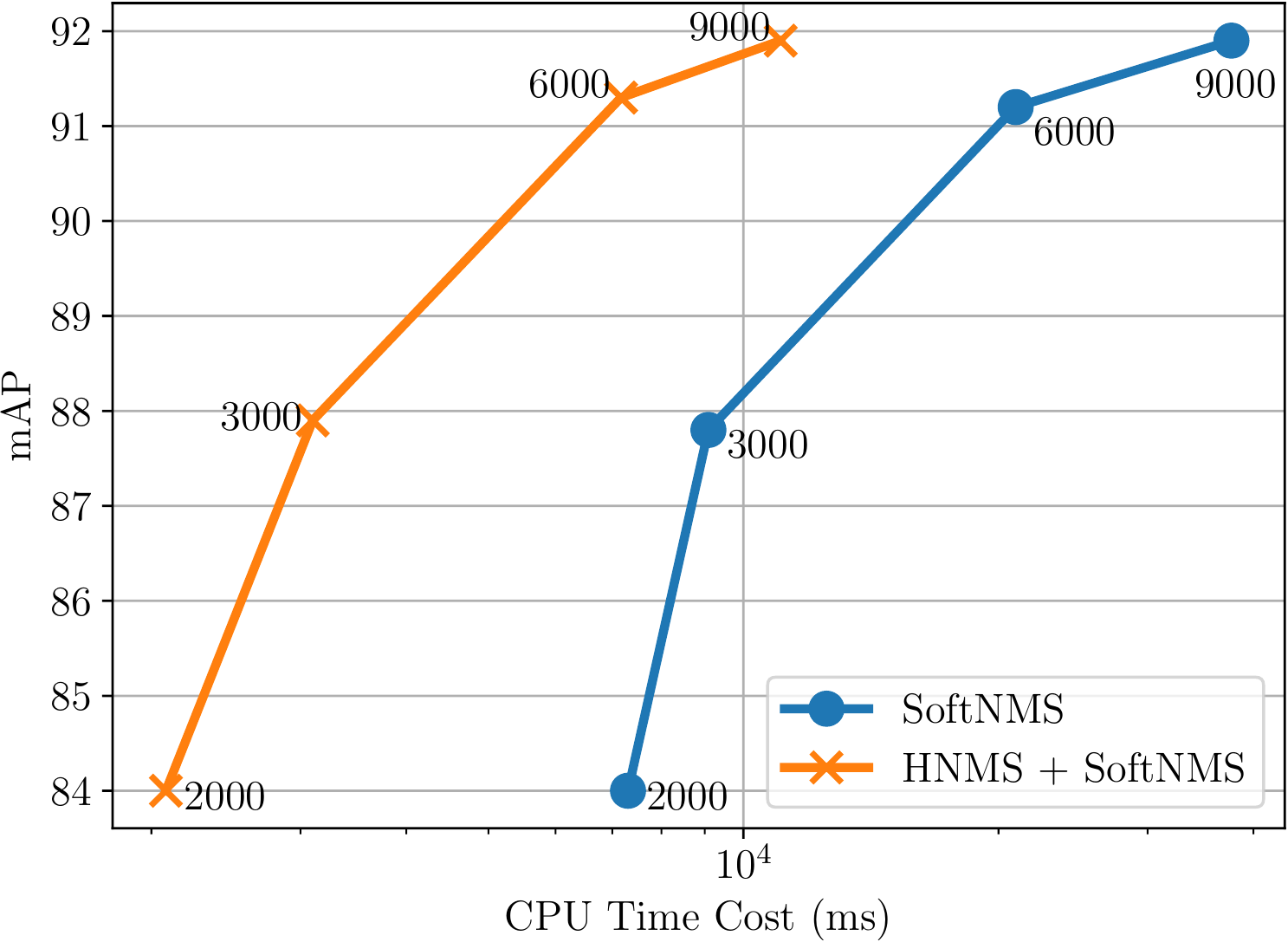} &
	\includegraphics[width=0.32\linewidth]{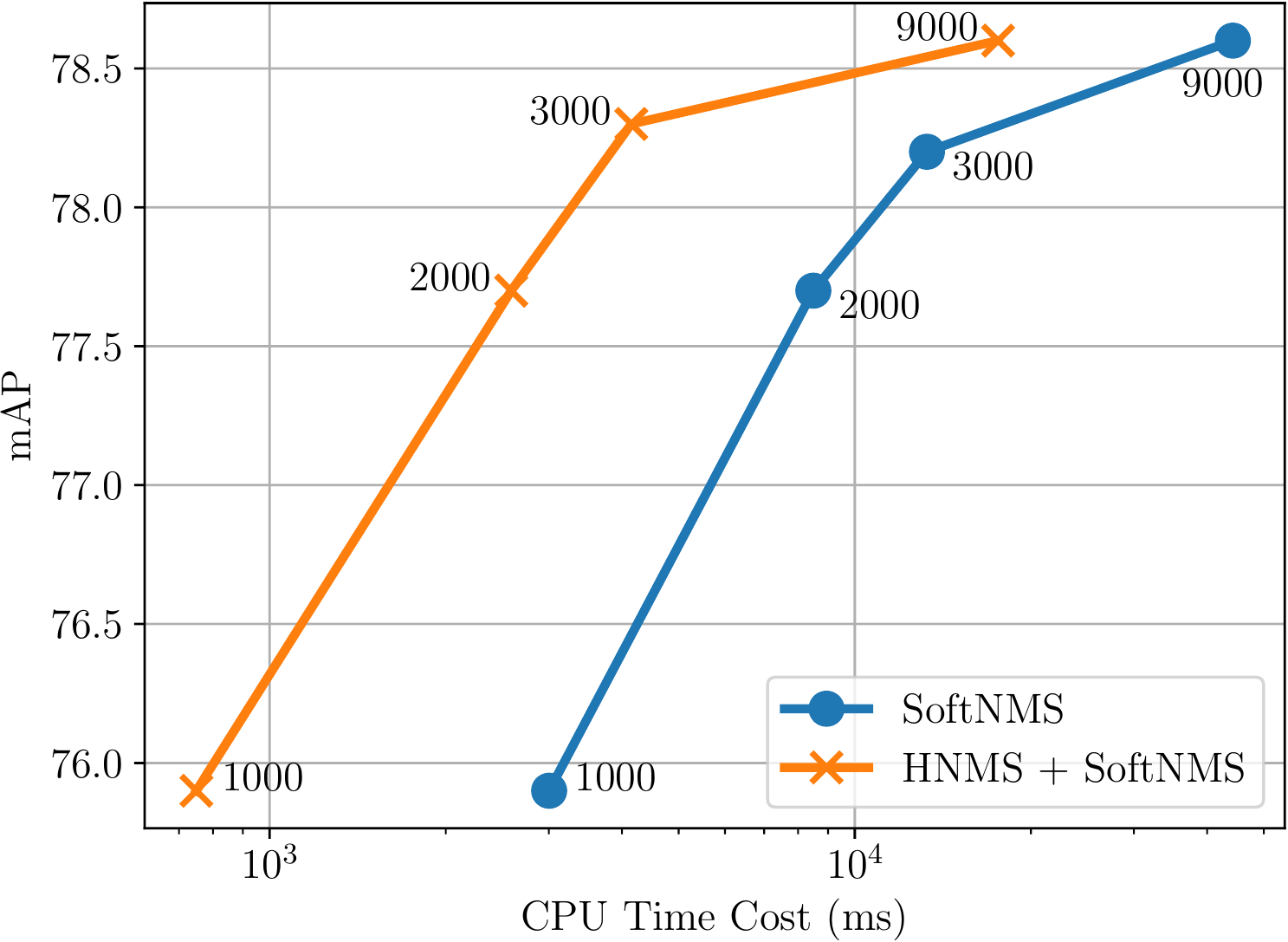} 
	\end{tabular}
\caption{mAP vs Time cost with different maximum numbers of boxes
for suppression filtering in Retina-R50-FPN. SoftNMS is
used and HNMS ($K=1$, $\alpha=0.73$) is as pre-filtering. }
\label{fig:softnms}
\end{figure}

\subsection{Application on SoftNMS}\label{sec:softnms}

One alternative to NMS is SoftNMS~\cite{BodlaSCD17} to achieve higher accuracy. 
Instead of discarding the boxes, SoftNMS 
decreases the confidence score. 
Specifically, the algorithm iteratively 1) finds the box with the highest confidence and insert it to the \textit{visited} list, 
2) decrease all \textit{un-visited} boxes' confidence based on the
IoU similarity with the box just inserted to the \textit{visited} list. 
Thus, the algorithm's complexity is $O(N^2)$ ($N$ is the number of boxes). 
Due to the lack of an efficient GPU implementation\footnote{The difficult part is that each iteration finds the box with maximum confidence, which relies on previous iteration.}, 
we mainly compare the time cost on CPU.

Table.~\ref{tbl:soft_retina} shows the experiment results with
Retina-R50-FPN on the three datasets with $P_1 = 9000$ (maximum nubmer of boxes used for NMS in each feature map). 
As can be seen, the time cost of SoftNMS is quite large, e.g. 40.7 seconds
on CARPK. By pre-filtering with the proposed
HNMS, the time cost can be significantly reduced, e.g. to 10.2 seconds on CARPK
($K=1$), with no accuracy regression.
With more hashing functions (larger $K$), the time cost 
can be further reduced since more boxes are pre-filtered.
One observation is that the time cost of the SoftNMS without pre-filtering is not consistent across the three datasets, because the confidence decreasing is skipped if the box has no overlap, which makes the time cost dependent on the data distribution.
Compared with NMS, SoftNMS improves the accuracy by 0.5 in CARPK, 0.3 in SKU-110K and 0.9 in CrowdHuamn.

\noindent\textbf{Varying $P_1$}. By altering different values of $P_1$, we arrive at Fig.~\ref{fig:softnms}, which clearly shows
the necessity of increasing $P_1$ and the significant gains with 
HNMS as pre-filtering.

\end{document}